\documentclass[conference]{IEEEtran}
\IEEEoverridecommandlockouts
\usepackage{cite}
\usepackage{amsmath,amssymb,amsfonts}
\usepackage{algorithmic}
\usepackage{graphicx}
\usepackage{textcomp}
\usepackage{xcolor}
\def\BibTeX{{\rm B\kern-.05em{\sc i\kern-.025em b}\kern-.08em
    T\kern-.1667em\lower.7ex\hbox{E}\kern-.125emX}}
    
\usepackage[colorinlistoftodos]{todonotes}

\usepackage{xspace}
\usepackage{comment}
\usepackage{listings}
\usepackage{multirow}
\usepackage{makecell}
\usepackage{subcaption}
\usepackage{enumerate}
\usepackage[ruled, vlined]{algorithm2e}
\usepackage{url}
\usepackage{flushend}

\newcommand{\name}{{\scshape Hawkeye}\xspace}
\iftrue 

\fi
\newtheorem{defn}{Definition}

\newcommand{\eg}{{\em e.g.,}\xspace}

\begin{document}

\title{\name: Adversarial Example Detector for Deep Neural Networks
}

\author{\IEEEauthorblockN{Jinkyu Koo}
\IEEEauthorblockA{\textit{Purdue University}\\
kooj@purdue.edu}
\and
\IEEEauthorblockN{Michael Roth}
\IEEEauthorblockA{\textit{Google}\\
roth28@purdue.edu}
\and
\IEEEauthorblockN{Saurabh Bagchi}
\IEEEauthorblockA{ \textit{Purdue University}\\
sbagchi@purdue.edu}
}

\maketitle

\begin{abstract}
Adversarial examples (AEs) are images that can mislead deep neural network (DNN) classifiers via introducing slight perturbations into original images. Recent work has shown that detecting AEs can be  more effective against AEs than preventing them from being generated. However, the state-of-the-art AE detection still shows a high false positive rate, thereby rejecting a considerable amount of normal images.
To address this issue, we propose \name, which is a separate neural network that analyzes the output layer of the DNN, and detects AEs.
\name's AE detector utilizes a quantized version of an input image as a reference, and is trained to distinguish the variation characteristics of the DNN output on an input image from the DNN output on its reference image.
We also show that cascading our AE detectors that are trained for different quantization step sizes can drastically reduce a false positive rate, while keeping a detection rate high.

\end{abstract}


\section{Introduction}

Image classification problems have achieved great success using deep neural networks (DNNs).
However, recent work has reported that DNN classifiers have a security issue such that
small perturbations to the input that humans may not recognize can drastically change their output \cite{Goodfellow2015,Kurakin2016a,Papernot_SP16}.
This security vulnerability is critical, since it means that for example, an adversary can make an autonomous vehicle mis-recognize a stop sign as a yield sign \cite{Papernot2017}.

Such perturbed inputs are called {\em adversarial examples (AEs)}. Adversaries can indeed make AEs with minimal perturbation, utilizing the gradient of the training cost function or the DNN model output (see Section \ref{sec:background}) \cite{Goodfellow2015,Kurakin2016b,Papernot2016}.
In order to defend against AEs, there have been many solutions proposed \cite{Gu14,Goodfellow2015,Papernot_SP16,Nayebi17}. Most of these defense mechanisms are modifying training methods or DNN architectures to hide the gradients near input data points so that adversaries get harder to generate AEs. However, these methods are not effective enough to prevent AEs from being generated, especially when the perturbation level is high (though still not recognizable to humans)
or when AEs are made using a substitute model with the same input and output formats, but with different architecture, initialization, or training methods, \textit{i.e.}, in so-called black-box attack methods \cite{Szegedy13,Papernot2017,openAI}.

Due to this limitation of existing defenses, recent work has turned to detecting AEs rather than making the DNN be robust against creating AEs \cite{corr/GrosseMP0M17,metzen2017detecting,corr/abs-1803-08533,ndss/Xu0Q18}. Detecting AEs is usually done by finding statistical outliers or training separate sub-networks that can distinguish between AEs and normal images. The state-of-the-art for detecting AEs, Feature Squeezing (FS)~\cite{ndss/Xu0Q18} can achieve a high detection rate (DR), but it comes at a high false positive rate (FPR), thereby rejecting a considerable fraction of benign images.

To reduce the FPR, while keeping a high DR, we propose \name, which is a separate neural network that detects AEs at the output layer of the application DNN. 
For a given input image $x$, \name analyzes the difference in the DNN outputs, denoted by $Z(x)=G(x)-G(x_q)$, where $G(x)$ is a DNN output when fed with $x$ and $G(x_q)$ is the one when fed with its reference image $x_q$. Here, the reference image is the one that is created from an input image to be insensitive to the perturbation noise added by adversaries.
Thus, whether or not an input image $x$ contains the perturbation noise, the reference image $x_q$ can stay almost the same. 
In order to create a reliable reference image, we apply quantization to an input image, \textit{i.e.}, our reference image is the quantized version of an input image.
Quantization is a widely used method to remove noise in image or signals in general~\cite{ndss/Xu0Q18,corr/LiangLSLSW17}. We co-opt it to detect AEs. 

When an AE denoted by $x^*$ enters into the DNN, we have $Z(x^*)= G(x^*)-G(x^*_q)$. Our key idea is that the vectors $Z(x)$ and $Z(x^*)$ will have very different patterns, for many possible $(x, x^*)$ combinations. The difference in patterns can be detected by our separate neural network that we called the AE detector or AED in short.
\name's AED is trained to distinguish such variations in $Z(x)$ from when the input $x$ is normal to when $x$ is an AE. We find that with the right choice of the quantization step size, we can nullify the malicious noise if any and have a good reference image, which, in turn, leads to good performance of the AED.

In \name, we also propose to use cascading AEDs that are trained using different quantization step sizes. Here, cascading means that we ask multiple AEDs and decide that an input image is an AE, only when all AEDs agree unanimously. Different step sizes may lead to different characteristics in $Z(x)$ (see Figure \ref{fig:corr}), and thus each AED based on a specific step size may learn something different about the same set of normal images and corresponding AEs.
Hence, cascading AEDs of various step sizes can increase our confidence when we raise a flag for a potential AE and thus further reduce FPR. 
We find empirically that training multiple AEDs with {\em different} quantization steps and even with the {\em same} training corpus reduces the correlation among them and cascading such AEDs reduces the FPR without degrading the DR.

We claim to make the following contributions here:
\begin{enumerate}
  \addtolength\itemindent{-0.5em}
  \addtolength\listparindent{-1.5em}
\item We create a novel way to detect an AE, which does not rely on any specific characteristic of the image or its modification. Rather it relies on creating a reference quantized image and passing both the original image and the quantized image through the same application DNN. We show that using a vector representation of the difference is highly effective in telling apart benign versus adversarial images. We use an appropriately trained neural network for detecting the difference. 

\item To further reduce FPR, we propose cascading AEDs that are trained for different quantization step sizes. We see that by intelligently choosing the different quantization steps, we can produce variants of AEDs, such that cascaded detection can significantly reduce an FPR compared to the FPR of an individual AED, while keeping a DR high.

\item We create our AED and evaluate on two widely used datasets, MNIST \cite{mnist} and ImageNet \cite{imagenet_cvpr09}. We show that compared to a recent state-of-the-art detector called Feature Squeezing (FS)~\cite{ndss/Xu0Q18}, \name achieves a much lower FPR (3.8\% in MNIST and 18\% in ImageNet) for the same high DR (near 100\%). Further, by cascading two detectors, our FPR is reduced by 20X in MNIST and by 2X in ImageNet. 

\end{enumerate}

Our AED is a separate, simple neural network that does not require any modification to the target DNN. 
Due to the difficulty of training an application DNN from scratch, this is often a desired characteristic.
Note that like other AE detection methods, \name is orthogonal to prior work whose focus is modifying a DNN to be robust against AEs by gradient masking. Thus, it can be used together with such defense mechanisms to achieve better protection.

This paper is organized as follows. Section \ref{sec:background} first reviews popular methods to generate AEs and the state-of-the-art AE detection scheme, introducing notations and concepts we use. Then, we propose our solution in Section \ref{sec:solution}. Experimental results are given in Section \ref{sec:exp}, and miscellaneous important things to consider are discussed in Section \ref{sec:discussion}.
In Section \ref{sec:related}, we describe the previous work that is related, but not directly overlapped to our work. Lastly, we conclude in Section \ref{sec:concusion}.

\section{Background}\label{sec:background}

\subsection{Classification problem}

\begin{figure}[t]
\centering
\includegraphics[width=0.9\linewidth]{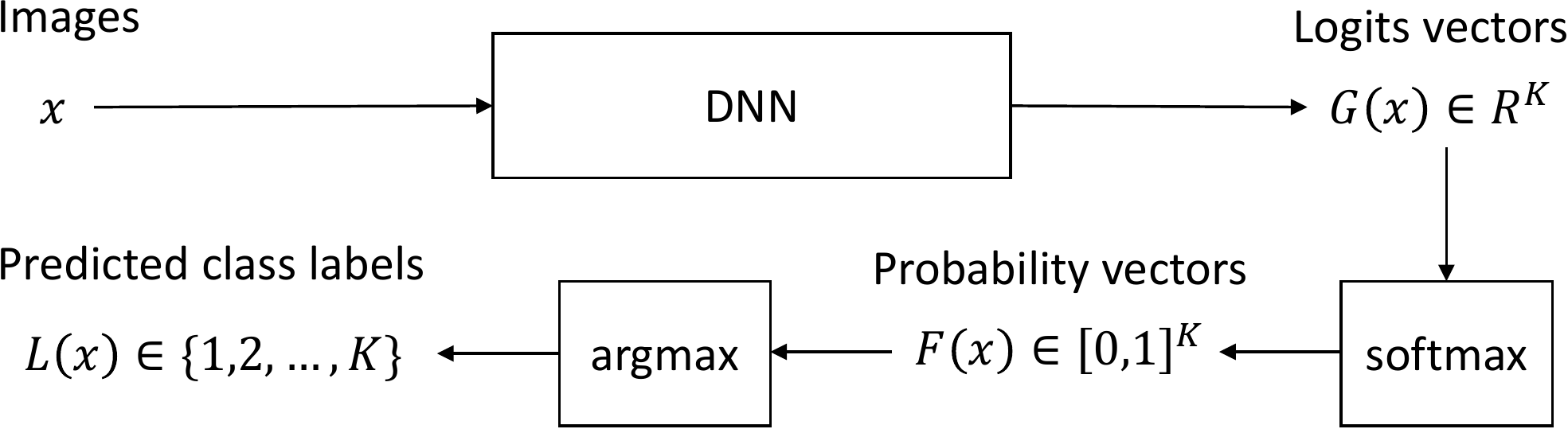}
\caption{A general framework of classification problems.}
\label{fig:classification}
\end{figure}	

We consider a classification problem using a DNN model as shown in Figure \ref{fig:classification}.
For an image $x$, the DNN computes a logits vector $G(x) \in R^K$ that is a non-normalized prediction, where $K$ denotes the number of classes.
Then, $G(x)$ is ordinarily passed to a softmax function that normalizes it, producing a probability vector $F(x) \in [0,1]^K$ that estimates the probability of each of the $K$ classes. The $F(x)$ is a model output trained by minimizing the average of the cost function $J(x,y)$ that computes the cross entropy as:
\begin{align}
J(x,y)=-\sum_{\forall i} [Y(x)]_i \log [F(x)]_i,
\end{align}
where $y \in \{1,2,\ldots,K\}$ is an integer representing the true class label of $x$, $Y(x)$ denotes a one-hot vector that has 1 at the index corresponding to $y$ and 0 otherwise, and $[V]_i$ means the $i$-th element of a vector $V$.
Thus, minimizing $J(x,y)$ means that we try to shape $F(x)$ in such a way that $[F(x)]_y > [F(x)]_j$ for any $j \ne y$.
The label predicted by a classifier is $L(x)$, which is the index of the largest probability, \textit{i.e.},
\begin{align}
L(x)=\arg\max_{i\in\{1,2,\ldots,K\}}{[F(x)]_i}.
\end{align}
When $L(x)$ matches the true index $y$, we say that the prediction is correct.

\subsection{Adversarial examples}


An adversarial example (AE), denoted by $x^{*}$, is one that alters a given input $x$ as:
\begin{align}
x^{*}=x+\delta,
\end{align}
where $\delta$ is the perturbation noise added by adversaries such that $L(x+\delta)\neq L(x)$.
That is, the adversarial example $x^{*}$ is the one that perturbs $x$ to mislead $x$'s class label.
In order not to be detected by human eyes or by simpler threshold-based detectors,
the magnitude of $\delta$ is typically chosen to be as small as possible.

\subsection{How to generate adversarial examples}\label{sec:perturbation}
Here, we describe representative methods to generate AEs, which we use in our experiments.\footnote{We also considered other recent methods, but they were not effective in generating AEs. See more detail in Section \ref{sec:selected_method}.}

\begin{itemize}
\item \textbf{Fast gradient sign method (FGSM)} by \cite{Goodfellow2015} perturbs all input pixels by the same quantity $\epsilon$ in the direction of a gradient sign, \textit{i.e.},
\begin{align}
x^{*} = x + \epsilon \text{ } \text{sign}\left(\frac{\partial J(x,y)}{\partial x} \right),
\label{eq:fgsm}
\end{align}
where sign$(v)$ is 1 if $v>0$, -1 if $v<0$, and 0 if $v=0$.
In other words, FGSM attempts to make an AE by adding a noise to each pixel in the direction that maximizes the increment in the cost function.
The value of $\epsilon$ is chosen to be a multiple of $\epsilon_0$, which corresponds to the magnitude of one-bit change in a pixel. We use the term {\em ``perturbation level''} for $\epsilon$.

\item \textbf{Iterative FGSM (I-FGSM)} by \cite{Kurakin2016a} iteratively applies FGSM (say, $N$ times) with the minimal amount of perturbation at a time as follows:
\begin{align}
x_{n+1} = \text{Clip}\left\{ x_{n} + \epsilon_0 \text{sign}\left(\frac{\partial J(x_{n},y)}{\partial x_{n}} \right)\right\},
\label{eq:ifgsm}
\end{align}
with $x_0=x$ and $x_N=x^{*}$. Here, Clip$\{\cdot\}$ denotes a pixel-wise clipping operation, which ensures that the pixel value stays in the $\epsilon$-vicinity of the original value, and in the valid range. The value of $N$ is heuristically set as $\min(\epsilon/\epsilon_0+4, 1.25\epsilon/\epsilon_0)$ in \cite{Kurakin2016a} to incur a similar level of total perturbation, compared to FGSM.
The I-FGSM usually creates AEs more successfully than FGSM for the same $\epsilon$ because it can fine-tune the perturbation in one iteration based on the result from the previous iteration.

\end{itemize}

\subsection{Adversarial example transferability}
It was shown in \cite{Szegedy13} that AEs crafted to mislead a DNN often also mislead a substitute model of the DNN (\textit{e.g.}, of different initialization or architecture). 
This property is called the adversarial example transferability, and means that it is possible to generate AEs using a substitute model and perform a misclassification attack on a machine learning system {\em without} access to the target model. Such an attack is referred to as a black-box attack. A recent work \cite{Papernot2017} demonstrated black-box attacks in the scenario where adversaries cannot even access the training dataset.

Thus, in our evaluation, we will consider two classes of attacks---the black-box attack and the white-box attack. They are mainly distinguished by where the adversarial examples are constructed, assuming that the training dataset is always accessible.
We formally define our distinction between them as follows:
\begin{defn}
\label{def:black-box}
\noindent \textbf{Black-box attacks} are the attacks that fool a target model by adversarial examples made on a substitute model. The adversaries do not know the internal parameters of the target DNN. However, using the same training data set, they can train their own DNN model so that they can construct the gradients of the target model with high similarity.
\end{defn}
\begin{defn}
\label{def:white-box}
\noindent \textbf{White-box attacks} are the attacks that	attempt to mislead the target DNN using the adversarial examples crafted on the target model itself. The adversaries are assumed to have access to the target DNN and thus be able to compute the gradients of the target.
\end{defn}

Most existing defenses that hide the gradient perform poorly against black-box attacks, although they are quite effective against white-box attacks \cite{openAI,Tramr17,Kurakin2016b,Papernot2017}. 
This is because a substitute model can have the decision boundary that is quite similar to that of the target model \cite{Florian17}. 
The adversary simply reconstructs the gradients using a substitute model, thus bypassing the gradient masking employed by such defenses. 

\subsection{Feature Squeezing: State-of-the-art for detecting adversarial examples}

\begin{figure}[t]
\centering
\includegraphics[width=0.9\linewidth]{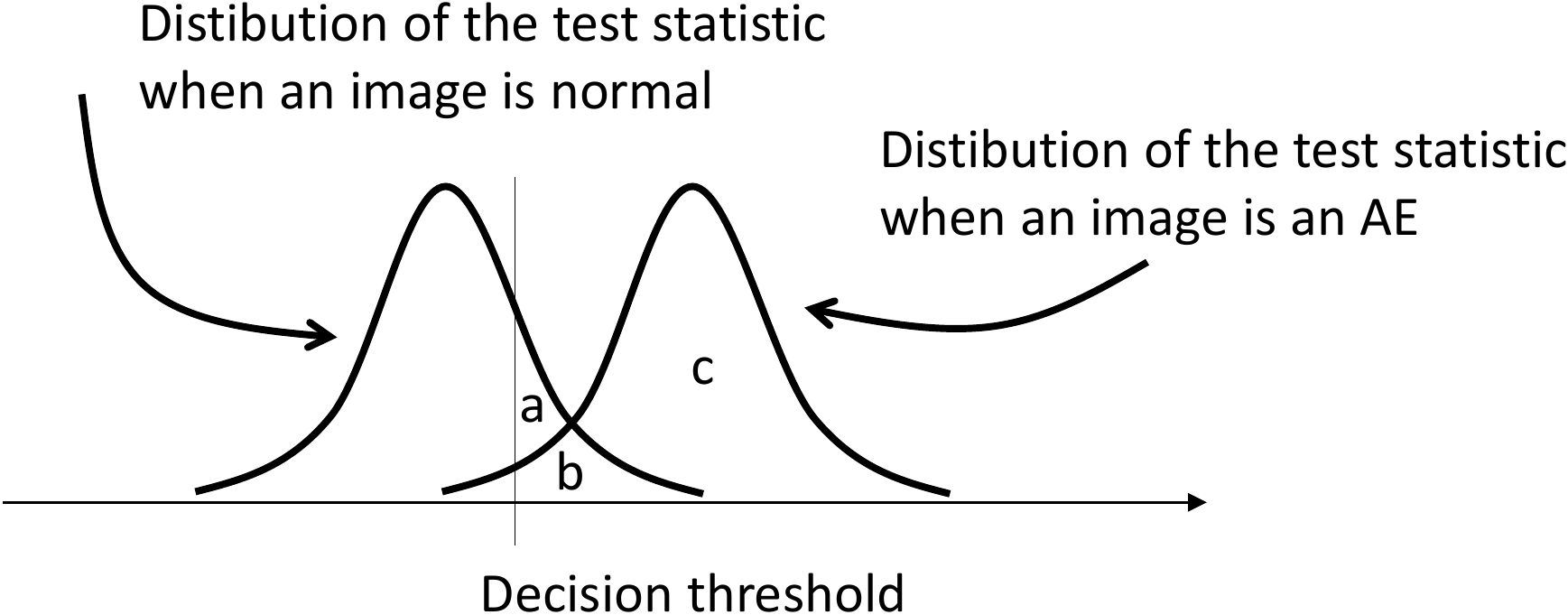}
\caption{General relationship between the detection rate and false positive rate. The detection rate is the sum of the areas $b$ and $c$, and the false positive rate is the sum of the areas $a$ and $b$. Reducing the decision threshold increases the detection rate, but the false positive rate also increases.}
\label{fig:dr_fpr}
\end{figure}	

The recent work, Feature Squeezing (FS) \cite{ndss/Xu0Q18} has proposed to use the squeezed input to detect AE. The squeezed input is the one with pixels of a reduced color bit depth, or the one that is spatially smoothed by blur filters, thereby being less sensitive to the perturbation noise added by adversaries.
FS computes two predictions --- one from the target DNN model with the original input and the second from the same DNN model with the squeezed input,
and then performs a threshold test on the $L_1$ distance between the two predictions.
If the $L_1$ distance is larger than a threshold, FS decides that the input image is an AE, otherwise it is benign.

{\em The intuition behind this design is that an adversarial modification will affect the original image but not the squeezed input image.}
We leverage this same insight. 
FS is reported to achieve a high DR for AEs, \eg the perfect DR for FPR = 0.05 for the MNIST dataset and DR = 0.64 for the same FPR for the ImageNet dataset.
However, its decision threshold needs to be fixed targeting a particular perturbation level. It performs poorly for perturbation levels that the threshold is not targeted for. In order to maintain the high DR for a wide range of perturbations, the value of the threshold can be chosen small enough, but this inevitably incurs a high FPR, as Figure \ref{fig:dr_fpr} explains.
For example, as we can see later in Figure \ref{fig:rates_white}, one threshold of FS can achieve a DR of 0.91 with an FPR of 0.038 at the perturbation level $\epsilon=32\epsilon_0$. However, this decision threshold gets a DR of 0.58 when the perturbation level goes down to $\epsilon=2\epsilon_0$. When we increase the DR at $\epsilon=2\epsilon_0$ to 0.90 by reducing the threshold, FS suffers from an FPR of 0.179. The high DR and low FPR values reported in the paper were obtained with large perturbation levels (which were left implicit in the paper), which may be detectable by a human-in-the-loop or simple threshold-based detectors.
Fundamentally, the drawback of FS is that there is a rigid mapping of the perturbation level used to generate the AE and the $L_1$ norm threshold and we show that using a richer detector can lead to more precise detection across a wide range of perturbation levels. 

\section{Solution approach}\label{sec:solution}

\begin{figure*}[t]
\centering
\includegraphics[width=0.9\linewidth]{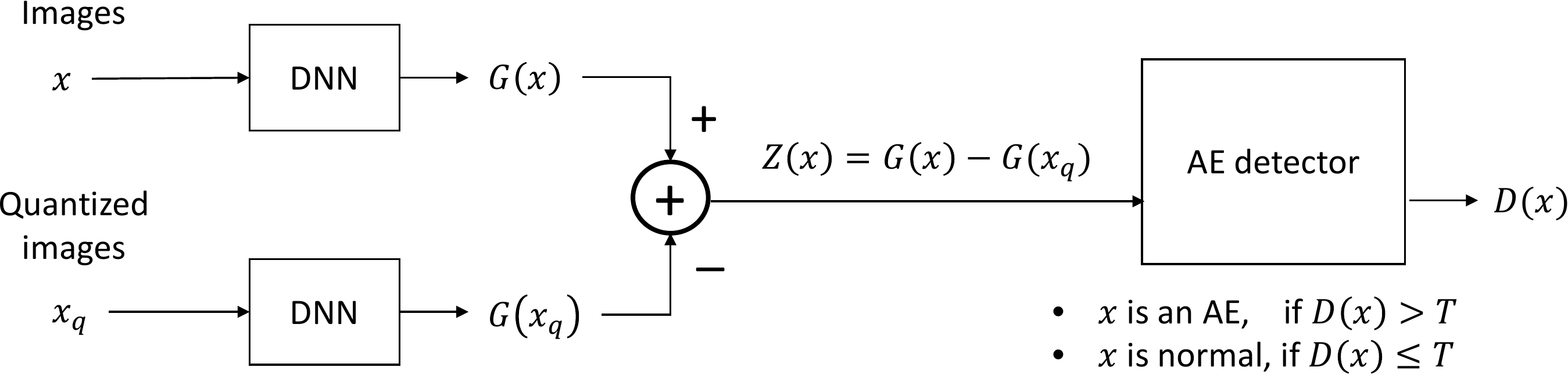}
\caption{Overview of \name's workflow.}
\label{fig:overview}
\end{figure*}	
\subsection{Reference inputs}
\label{sec:reference}
\begin{figure}[t]
\centering
\begin{subfigure}{0.9\linewidth}
\includegraphics[width=\linewidth]{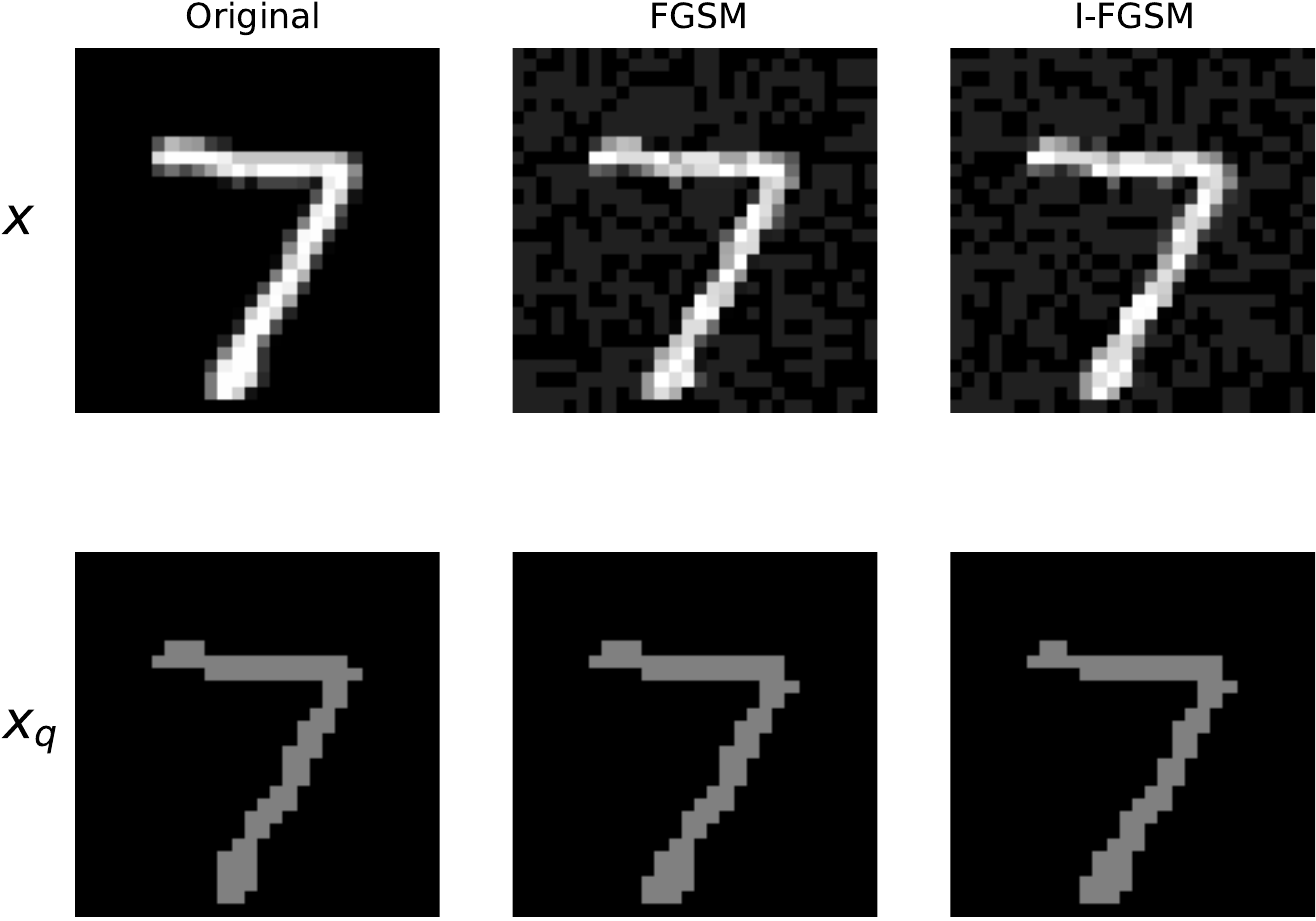}	
\caption{MNIST \cite{mnist} when $\epsilon=32\epsilon_0$ and $s=128$.} \label{fig:plot_example_mnist}
\end{subfigure}	
\\
\begin{subfigure}{0.9\linewidth}
\includegraphics[width=\linewidth]{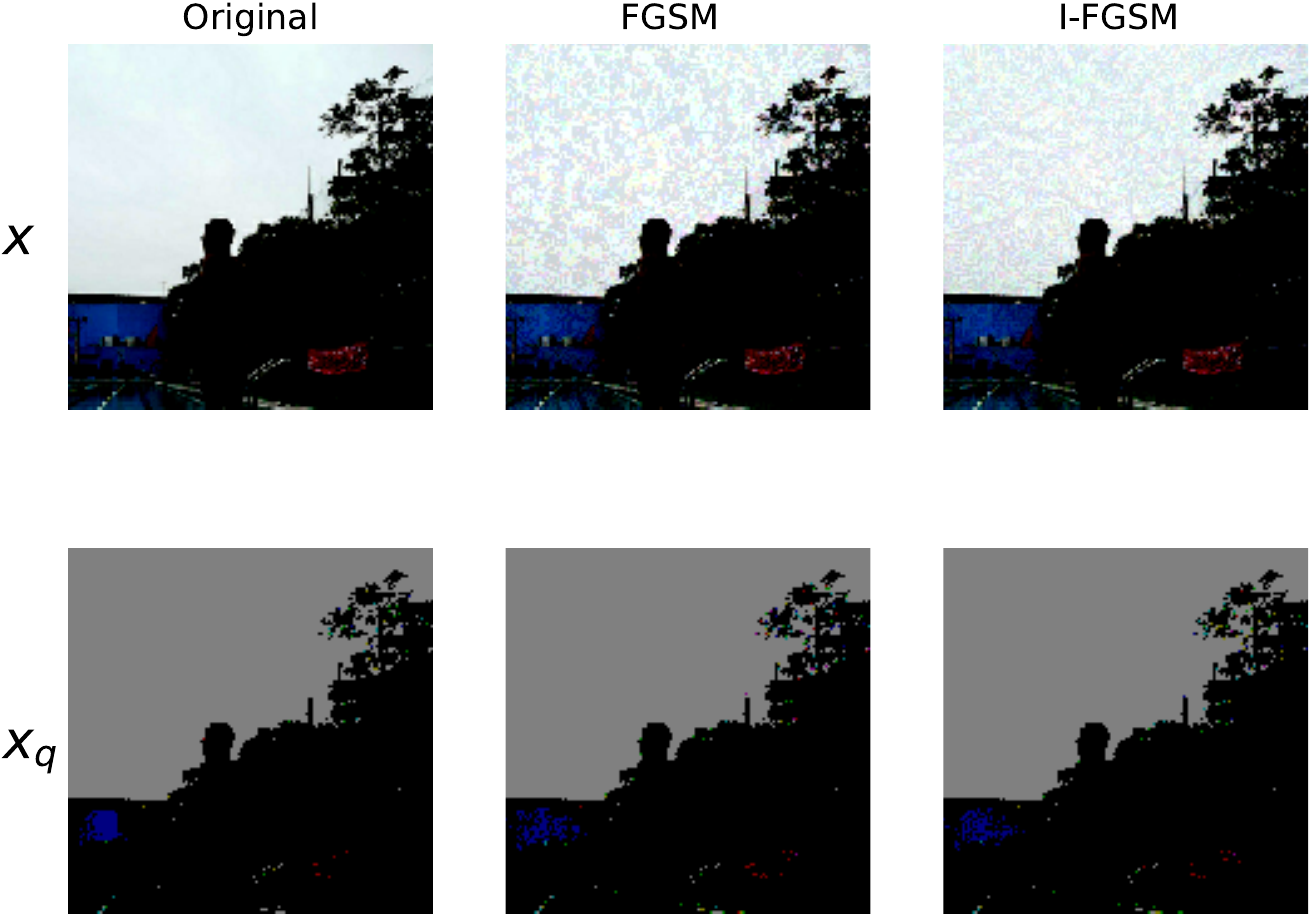}
\caption{ImageNet \cite{imagenet_cvpr09} when $\epsilon=8\epsilon_0$ and $s=64$.} \label{fig:plot_example_imagenet}
\end{subfigure}	
\caption{An example of quantization on the input image. The second column is the quantized version of the image in the first column. We can see that after quantization, a normal image and its corresponding AE become more similar to each other.}
\label{fig:example}
\end{figure}

Figure \ref{fig:overview} shows the overall workflow of \name. For a given image $x$, we consider a quantized image $x_q$, which is made by quantizing each pixel of $x$ with step size $s$. Typically, a pixel of an image is represented in 8 bits, ranging in value over $[0,255]$. Thus, quantizing with step size $s=128$, for example, means that a pixel of $x_q$ is represented as either 0 or 128, which is of 1-bit information.
In general, after quantization with step size $s$, a pixel value $v$ is represented as $s\lfloor v/s \rfloor$, where  $\lfloor \cdot \rfloor$ denotes a floor function.
The motivation to use such a quantized input is that as Figure \ref{fig:example} exemplifies, normal image and its corresponding AE become more similar after quantization, since the quantization process may nullify the noise added by adversaries that is smaller than the quantization step size $s$.\footnote{We also tested the smoothed input applying blur filters to see if it can act as a reference, as introduced in the FS paper. However, its performance was far worse than that with quantization. Thus we omit those results in our paper.} Thus, the quantized input $x_q$ can play a role as an invariant reference of an image, regardless of the existence of malicious additive noise.

\subsection{Adversarial example detector}
\label{sec:aed}
\begin{table}[t]
\caption{
AE detector architectures in MNIST and ImageNet. Here, FC denote the fully connected layer, where neurons  have connections to all activations in the previous layer, as seen in regular neural networks.
}
\label{table:arch}
\begin{tabular}{ccccc}
\hline
Layer number & Layer type & Activation & \begin{tabular}[c]{@{}c@{}}Output shape\\ (MNIST)\end{tabular} & \begin{tabular}[c]{@{}c@{}}Output shape\\ (ImageNet)\end{tabular} \\ \hline
Input        &            &            & 10                                                             & 1,000                                                             \\ \hline
1            & FC         & Relu       & 10                                                             & 1,000                                                             \\ \hline
2            & FC         & Relu       & 10                                                             & 100                                                               \\ \hline
3            & FC         & Relu       & 10                                                             & 10                                                                \\ \hline
4            & FC         & Sigmoid    & 1                                                              & 1                                                                 \\ \hline
\end{tabular}
\end{table}

To utilize $x_q$ as a reference, we compute $G(x)$ and $G(x_q)$, which are the logits vectors of the same DNN for the inputs $x$ and $x_q$, respectively, and take a difference between the two as $Z(x)=G(x)-G(x_q)$. This difference vector $Z(x)$ of length $K$ (\textit{i.e.}, the number of classes) is the input to our AED. The AED is a separate neural network, consisting of fully connected layers and making the detector output $D(x) \in [0,1]$ by a sigmoid activation function. Table \ref{table:arch} shows the detail of AED architecture. We train the AED in such a way that the detector output $D(x) $ becomes high if $x$ is an AE, and $D(x)$ becomes low if $x$ is normal. Then, we perform a threshold test seeing if $D(x)>T$, where $T$ is our threshold, and if this holds true, we mark the image as an AE.
An important design point is that the AED's neural network is far simpler than the application DNN we are trying to protect. Thus, the design and the training effort for the AED are minimal. 

In the detailed look at the AED, we maximize the average of the following cost function $J_d(x)$ to train it:
\begin{align}
J_d(x)=\left( 1-D(x) \right)+D(x^{*}),
\label{eq:detector_cost}
\end{align}
where $x^{*}$ is defined as in \eqref{eq:fgsm} with the value of the perturbation level $\epsilon$ that is randomly varying over the interval $[0, \epsilon_{max}]$.
That is, for each input $x$, we create $x^{*}$, feed $Z(x)$ and $Z(x^*)$ to the AED to form \eqref{eq:detector_cost}, and maximize it.
Here, $\epsilon_{max}$ is the perturbation level beyond which the AE is far too distorted from the original image and thus is easily recognizable by humans. Since $x^{*}$ is a potential AE crafted from $x$, what we want ideally is that $D(x^{*})=1$ and $D(x)=0$. Training the detector by maximizing the average of the cost function $J_d(x)$ means adjusting the internal weight parameters of the detector neural network towards achieving this ideal goal. Of course, the value of $D(x)$ seldom becomes exactly 0 or 1 in tests, regardless of whether $x$ is a normal image or an AE. Rather, $D(x)$ lies in between 0 and 1. Thus, we use a threshold $T$ (setting $T=0.5$ by default) and decide that $x$ is an AE only if $D(x)>T$.
Thus, the value of $T$ acts as a knob to control the trade-off between DR and FPR.
A high value of $T$ decreases FPR at the cost of lowering DR. 

Note that the quantized input is equivalent to one of the squeezed inputs that uses the reduced bit depth in FS \cite{ndss/Xu0Q18}. However, FS considers only the $L_1$ norm of $Z'(x)=F(x)-F(x_q)$ for threshold testing, thereby using a single scalar value and losing information relative to our design of taking the difference vector. Taking the difference after softmax normalization, \textit{i.e.}, $F(x)-F(x_q)$ instead of $G(x)-G(x_q)$, also loses information and reduces the detectability of FS.
In contrast, we use the whole of $Z(x)$, and analyze the subtle variation across all elements of $Z(x)$ in the AED to flag an image as adversarial or benign.

In order to train an AED, we use the same training images as were used for training the application DNN. For each such data point (which is by definition benign---call this as $x$), we create a set of AEs varying the perturbation level (using the standard adversarial example generation techniques described in Section~\ref{sec:perturbation}---call this the set of $x^*$'s). We now train the AED maximizing the cost function given in \eqref{eq:detector_cost}. 

It is also worth noting that our AED is a totally separated neural network that is attached to the output layer of the target DNN. That is, we do not require to modify the target DNN itself. This is often a desired characteristic, since in many use cases, transfer learning is used, which utilizes a DNN model that is already trained by someone else. This has become popular due to the difficulty of training a DNN from scratch.

\subsection{Cascading AE detectors of various quantization step sizes}

\begin{figure}[t]
\centering
\begin{subfigure}{0.45\linewidth}
\includegraphics[width=\linewidth]{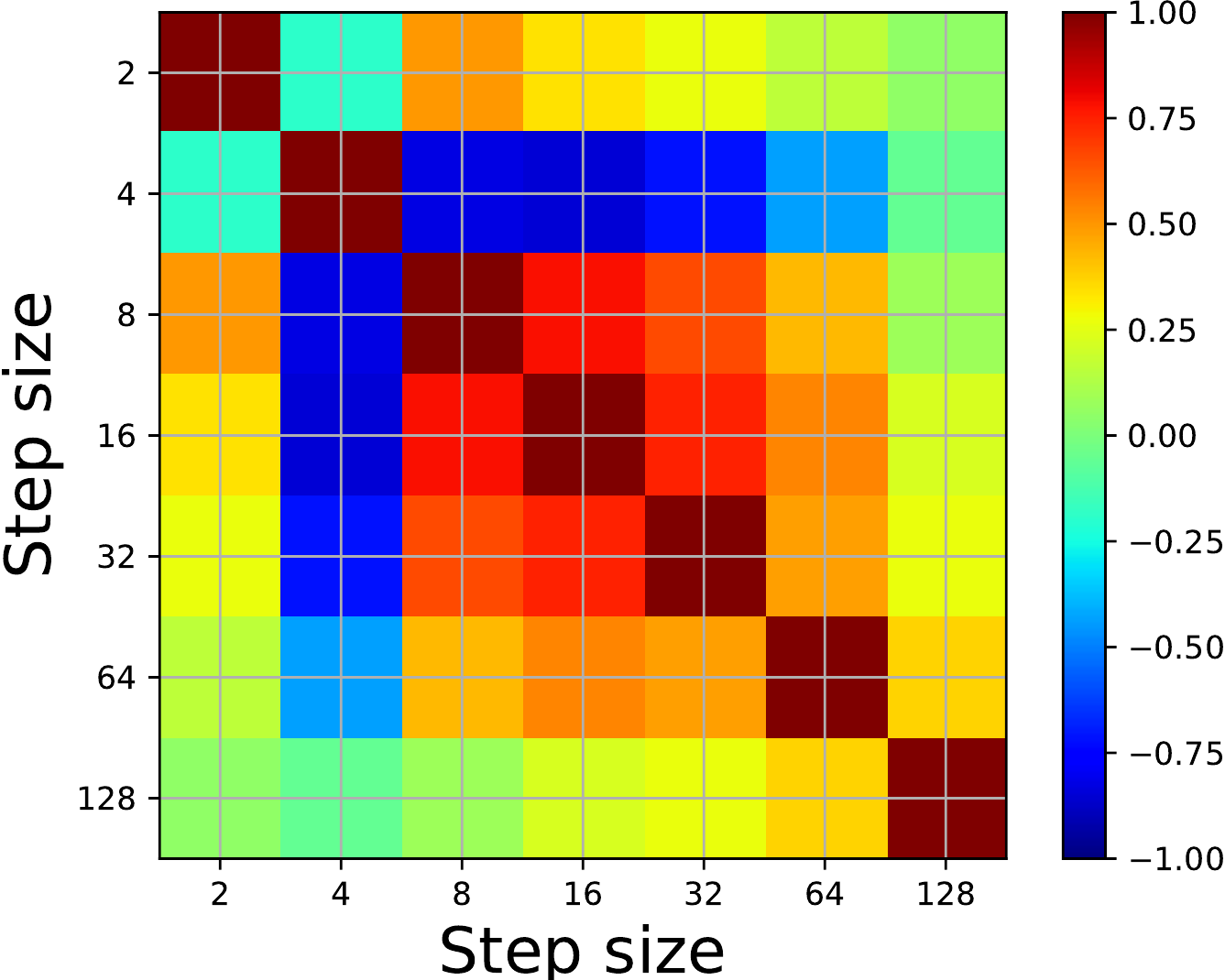}	
\caption{$\epsilon=2\epsilon_0$.} \label{fig:corr_2}
\end{subfigure}	
\hfil 
\begin{subfigure}{0.45\linewidth}
\includegraphics[width=\linewidth]{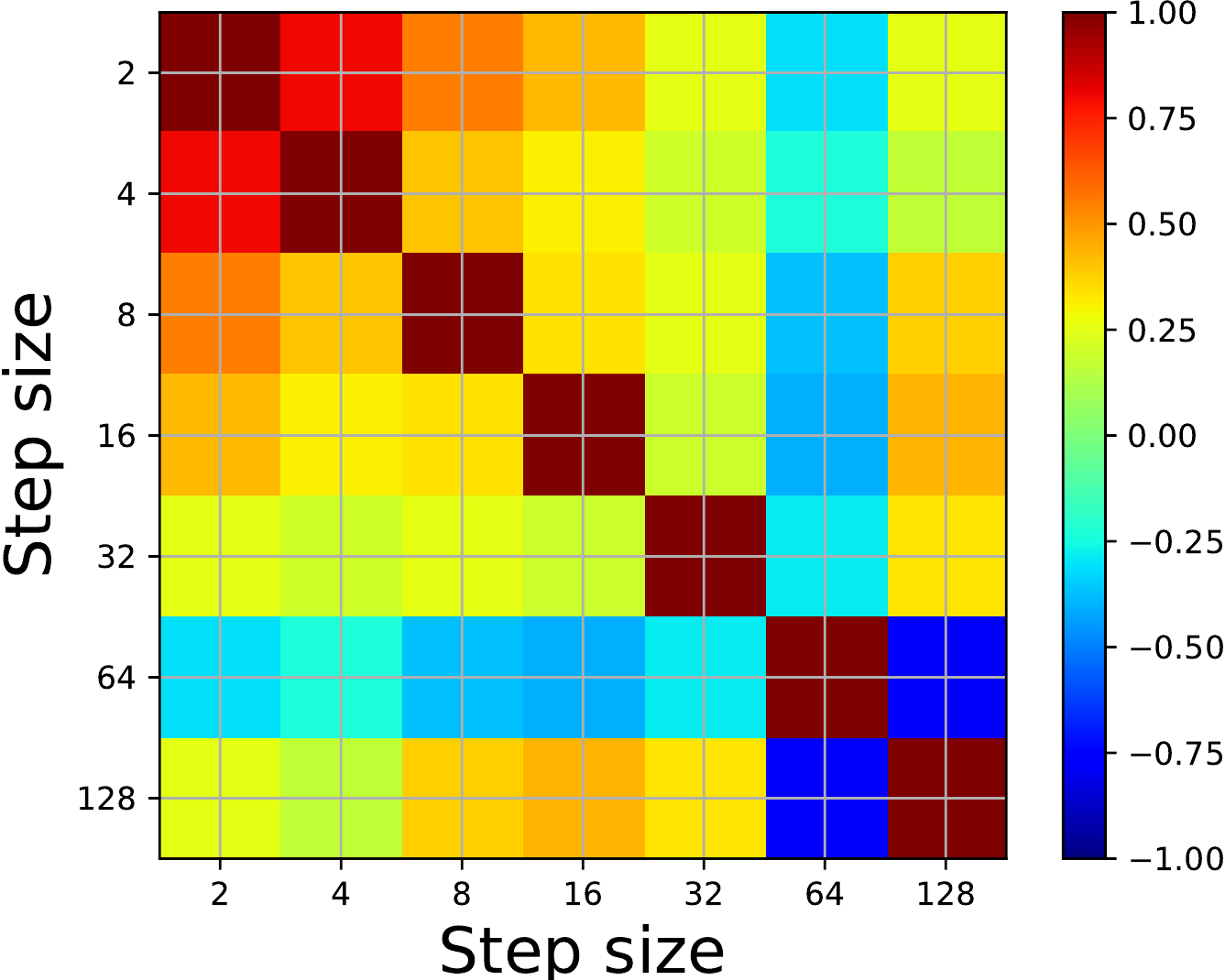}
\caption{$\epsilon=32\epsilon_0$.} \label{fig:corr_32}
\end{subfigure}	
\caption{Heatmaps showing Pearson correlation coefficients (PCCs) between $Z(x^*)$'s of different quantization step sizes in MNIST dataset \cite{mnist}. The value of a PCC is an average of 10,000 images, and $x^*$ is generated by \eqref{eq:fgsm}. We can see that the PCC between different step sizes is usually much lower than 1, meaning that the same $x^*$ may lead to a different $Z(x^*)$ for a different step size.}
\label{fig:corr}
\end{figure}

\begin{figure}[t]
\centering
\begin{subfigure}{0.6\linewidth}
\includegraphics[width=\linewidth]{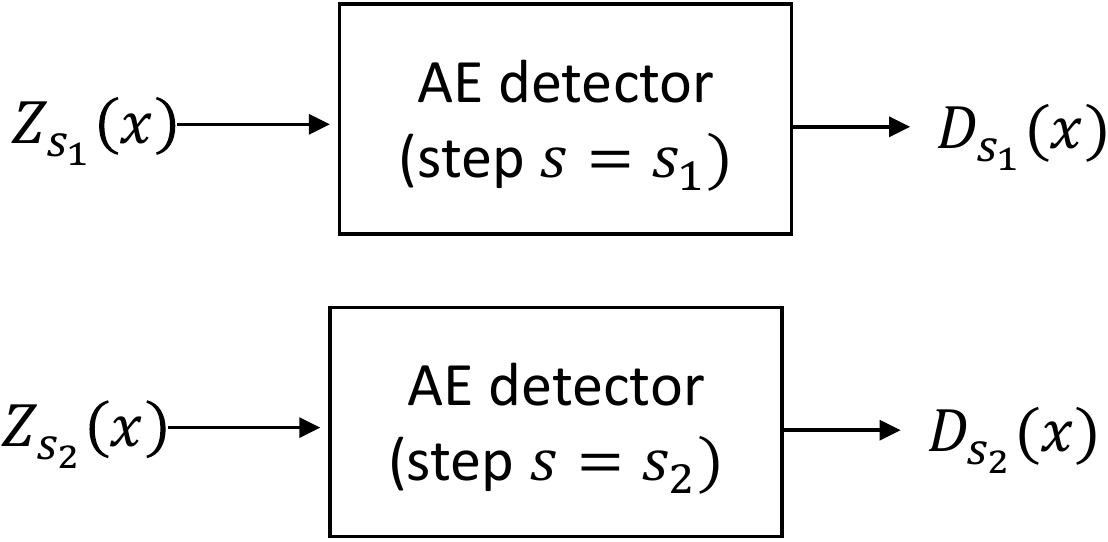}	
\caption{Adversarial example detectors are trained with different values of the quantization step $s$.} \label{fig:cascade_1}
\end{subfigure}	
\\\hfil 
\begin{subfigure}{0.6\linewidth}
\includegraphics[width=\linewidth]{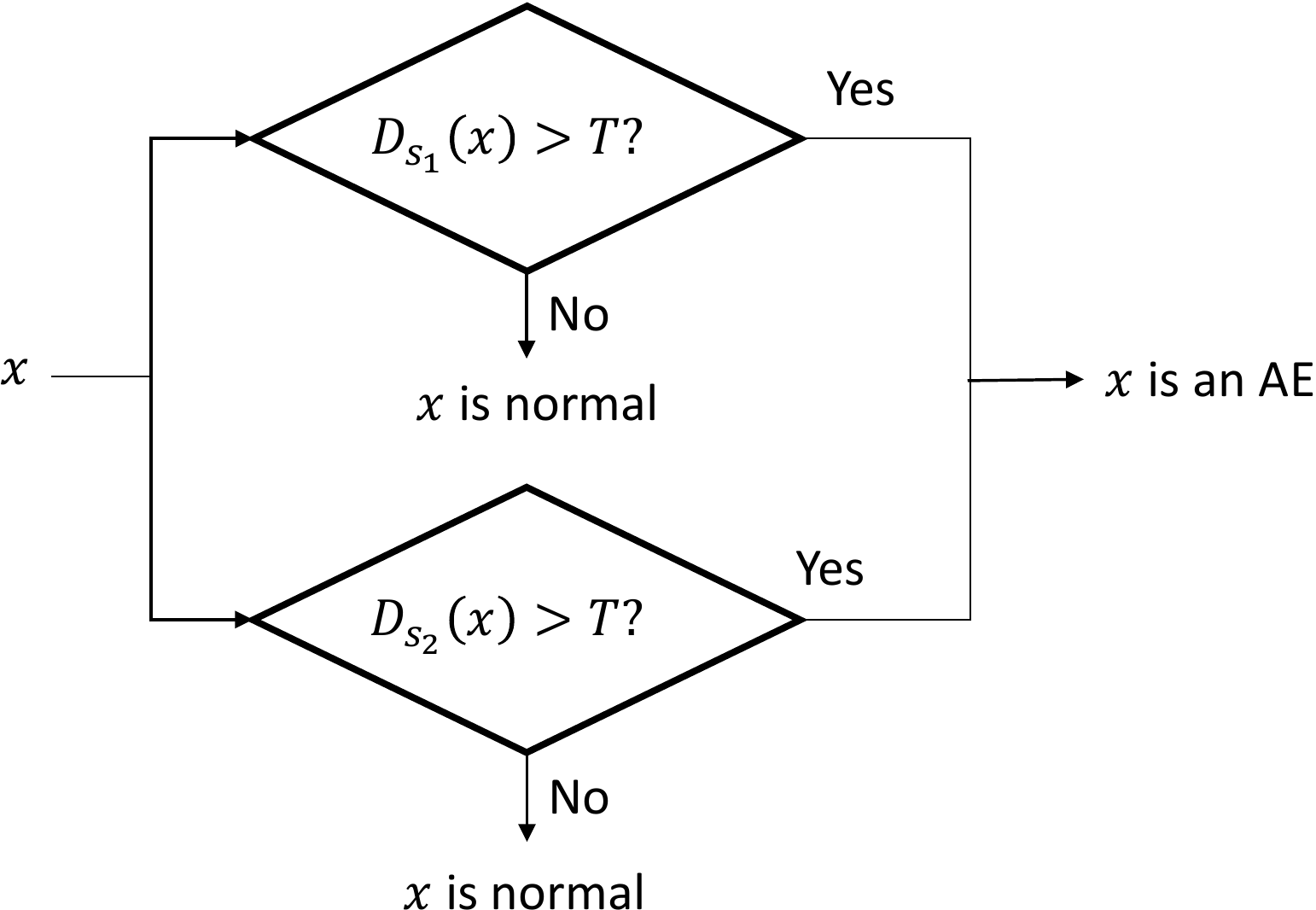}
\caption{An example of cascading two detectors of different values for the step $s$.} \label{fig:cascade_2}
\end{subfigure}	
\caption{Cascading detectors that use $s_1$ and $s_2$ as the quantization step size.}
\label{fig:cascade}
\end{figure}

Even for the same AE $x^*$, characteristics on $Z(x^*)$ may differ depending on the quantization step size $s$. Figure \ref{fig:corr} shows how $Z(x^*)$ can be different in terms of Pearson correlation coefficients (PCCs) according to the step sizes in MNIST dataset \cite{mnist}. In general, the PCC between two vectors becomes 1 when two vectors are perfectly correlated and 0 when there is no correlation. When PCC$=-1$, it means that the two vectors are perfectly correlated with the opposite sign. Let us denote $Z(x^*)$ for the step size $s$ by $Z_s(x^*)$. Then, in our case, any positive PCC lower than 1 between $Z_{s_1}(x^*)$ and $Z_{s_2}(x^*)$ implies that $Z_{s_1}(x^*)$ and $Z_{s_2}(x^*)$ may look different correspondingly to the magnitude of the PCC. Note that negative PCCs also mean that $Z_{s_1}(x^*)$ and $Z_{s_2}(x^*)$ are different enough, since they are the inputs to the AED, which is a non-linear system made by a neural network, where inputs of opposite sign typically result in completely different outcomes.
Thus, Figure \ref{fig:corr} results, where the PCC between different step sizes is usually much lower than 1, suggest to us that for the same $x^*$, we can have various forms of $Z(x^*)$ as the input to our AED by changing the step size.

Using such a characteristic of $Z(x^*)$, we can consider multiple variations of the AED, each of which uses a different step size, in order to decrease FPR of the overall \name system.
All variations of the AED using a different step size are trained experiencing a different form of $Z(x^*)$.
A cascade of AEDs is used {\em in parallel}. Only when every AED in the cascade designates an input as adversarial does \name flag it as an AE. 

Figure \ref{fig:cascade} shows an example of cascading two AEDs that use $s_1$ and $s_2$ as the quantization step sizes. Each detector is separately trained using its own step size, on the same $x$'s and correspondingly generated $x^*$'s. Then, any given image is double-checked by each of detectors and it is only classified as an AE when both detectors think it is an AE.
Note that in this case, we have the following if the AEDs are completely independent:
\begin{align}
\text{DR} &= (\text{DR of detector 1}) \times (\text{DR of detector 2}), \nonumber\\
\text{FPR} &= (\text{FPR of detector 1}) \times (\text{FPR of detector 2}). \nonumber
\end{align}
That is, when we cascade detectors, the total FPR is the product of FPRs of all detectors, and thus it goes down. However, the total DR is also the product of DRs of all detectors. Thus, unless each detector achieves DR $=1$, the total DR also decreases. For this reason, we only cascade detectors that have a DR higher than 0.99.
In practice, the lack of correlation among the different AEDs in a cascade helps us achieve this high DR, low FPR goal. 

\section{Experiment}\label{sec:exp}

In the section, we first introduce what datasets and DNN models are used, and see how successfully AEs can be generated in the models. Then,  until Section \ref{sec:eval_cascading}, we evaluate the performance of \name in comparison to FS against white-box attacks. In Section \ref{sec:black-box}, we see the performance against black-box attacks. In Section \ref{sec:effect}, we also see how well \name performs against other types of distorted images that can be generated by environmental effects on a camera \cite{corr/PeiCYJ17}.

\subsection{Datasets}

In our evaluations, we consider two popular datasets for image classification tasks, MNIST \cite{mnist} and ImageNet \cite{imagenet_cvpr09}. MNIST is a database of handwritten digit images of 28$\times$28 pixels in one channel, and labels are integers from 0 to 9, (thus $K=10$). It has a training set of 60,000 images, and a test set of 10,000 examples. ImageNet is a large visual database of over 14 million color images. Classification task is to choose one out of 1,000 labels. Thus, the value of $K$ is 1,000 here. We use its images after preprocessing to 224$\times$224 pixels in three channels via resizing and cropping. We use the ILSVRC 2012 curation \cite{imagenet_2012} of ImageNet that has 1.28 million images for training and 50 thousand images, which are not overlapped to the training images, for test purpose.

\subsection{DNN models}\label{sec:dnn_models}
\begin{table}[t]
\caption{
DNN models used for each dataset in evaluations. Accuracy here is calculated over the test portion of each dataset.
}
\label{table:model}
\begin{tabular}{cccc}
\hline
Dataset  & Model                                & Top-1 accuracy & Top-5 accuracy \\\hline
MNIST    & 4-layer CNN \cite{cnn_mnist} &  0.99          & -              \\\hline
ImageNet & ResNet-18 \cite{resnet}   &  0.66          &  0.84          \\
\hline
\end{tabular}
\end{table}

We set up a DNN model for each dataset that can achieve the state-of-the-art performance. For MNIST, we use a simple convolutional neural network (CNN) that is a model given in the Tensorflow tutorial \cite{cnn_mnist}. It consists of two convolutional layers and two fully connected layers.  Training is done for 5 epochs using Adam optimizer \cite{DBLP:journals/corr/KingmaB14} with learning rate = 0.0002.
For ImageNet, we use a Residual Network of 18 layers (ResNet-18) \cite{resnet}. The image is resized with its shorter side randomly sampled in $[256, 480]$ for scale augmentation, and then a 224$\times$224 crop is randomly sampled from an image or its horizontal flip. We train the ResNet-18 model for 20 epochs using Nesterov optimizer \cite{Nesterov:2014:ILC:2670022} with momentum$=0.9$. The learning rate starts from 0.1 and is divided by 10 when the error plateaus.
In both datasets, we use a mini-batch of size 100.
The classification performance of the DNN models is shown in Table \ref{table:model}, where the top-$k$ accuracy means the rate that one of top $k$ predictions matches the true label.
All the rates in our paper, including the accuracies here,  are calculated using the test portion of each dataset.

\subsection{Attack success rates without a defense}\label{sec:selected_method}

\begin{figure}[t]
\centering
\begin{subfigure}{0.70\linewidth}
\includegraphics[width=\linewidth]{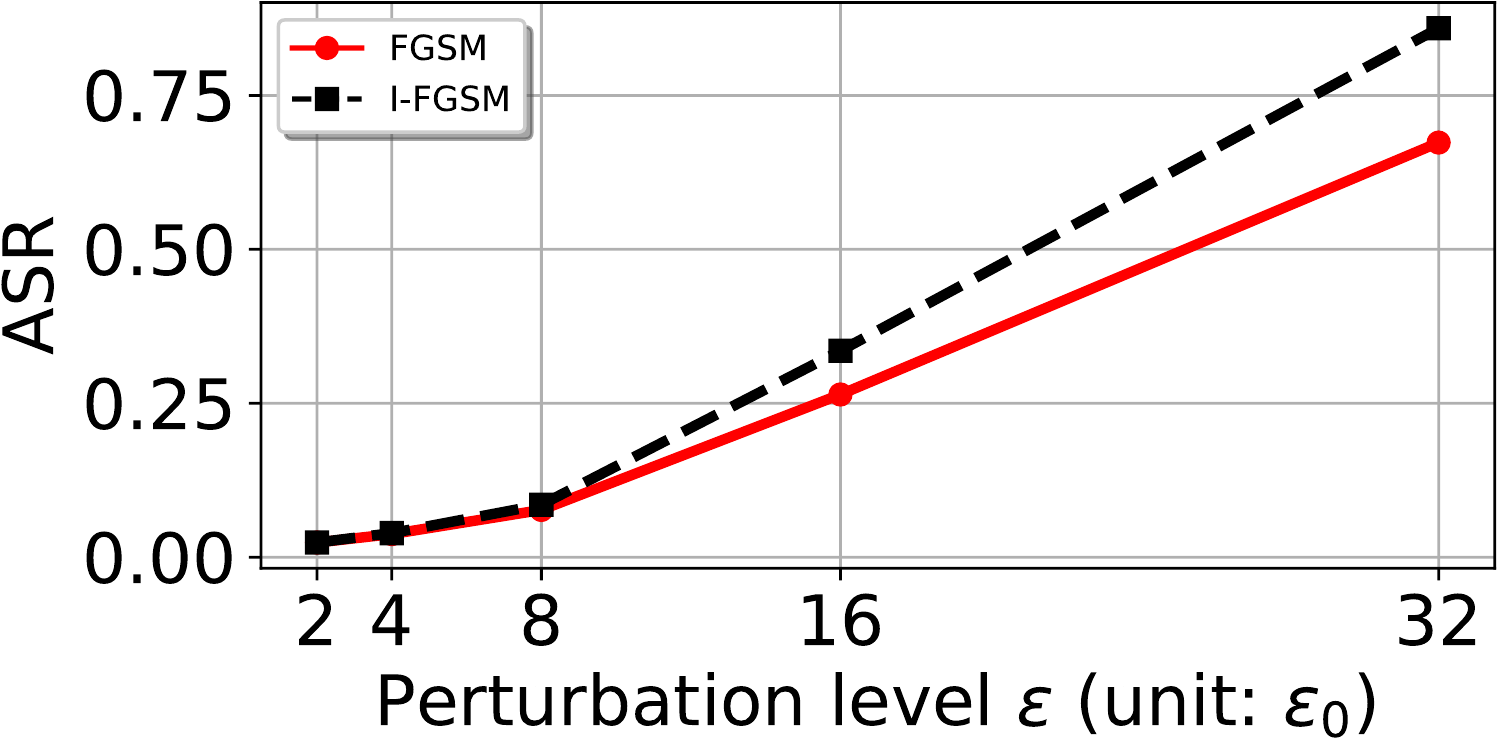}	
\caption{MNIST.} \label{fig:success_mnist}
\end{subfigure}	
\\
\begin{subfigure}{0.70\linewidth}
\includegraphics[width=\linewidth]{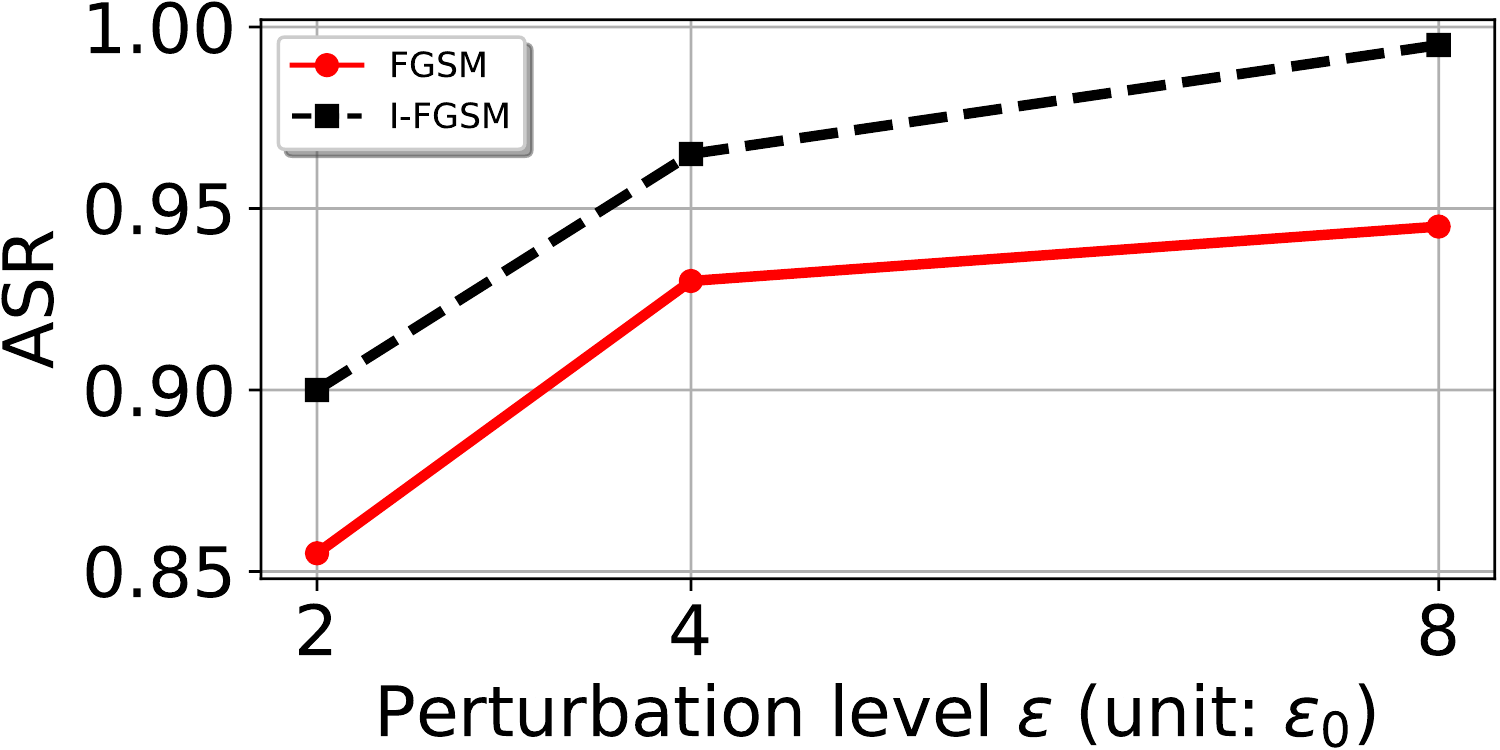}
\caption{ImageNet.} \label{fig:success_imagenet}
\end{subfigure}	
\caption{Attack success rate (ASR) at a given perturbation level $\epsilon$, when no defense mechanism is employed.}
\label{fig:success}
\end{figure}

\begin{table*}[t]
\centering
\caption{
Detector performance in ``(detection rate, false positive rate)" according to the perturbation level $\epsilon$ (unit: $\epsilon_0$). AEs are generated using FGSM.
}
\label{table:detector_step}
\begin{subtable}{0.99\textwidth}
\centering
\caption{MNIST.}
\label{table:detector_step_mnist}
\begin{tabular}{cccccccc}
\hline
\multirow{2}{*}{$\epsilon$} & \multicolumn{7}{c}{Step size $s$}                                                                                               \\ \cline{2-8} 
                         & 2               & 4              & 8              & 16              & 32             & 64             & 128            \\ \hline
2                        & (0.092, 0.048)  & \textbf{(1.000, 0.093)} & (0.866, 0.082) & (0.921, 0.026)  & (0.519, 0.022) & (0.167, 0.038) & (0.113, 0.051) \\ \hline
4                        & (0.194, 0.048) & (0.218, 0.093) & \textbf{(0.995, 0.082)} & (0.984, 0.026)  & (0.944, 0.022) & (0.628, 0.038) & (0.266, 0.051) \\ \hline
8                        & (0.189, 0.048)  & (0.188, 0.093) & (0.089, 0.082) & \textbf{(1.000, 0.026)}  & \textbf{(1.000, 0.022)} & (0.982, 0.038) & (0.706, 0.051) \\ \hline
16                       & (0.181, 0.048) & (0.181, 0.093) & (0.095, 0.082) & (0.072, 0.026) & \textbf{(1.000, 0.022)} & \textbf{(1.000, 0.038)} & (0.994, 0.051) \\ \hline
32                       & (0.232, 0.048)  & (0.198, 0.093) & (0.070, 0.082) & (0.047, 0.026)  & (0.018, 0.022) & \textbf{(1.000, 0.038)} & \textbf{(1.000, 0.051)} \\ \hline
\end{tabular}
\end{subtable}
\\
\vspace{5pt}
\begin{subtable}{0.99\textwidth}
\centering
\caption{ImageNet.}
\label{table:detector_step_imagenet}
\begin{tabular}{cccccccc}
\hline
\multirow{2}{*}{$\epsilon$} & \multicolumn{7}{c}{Step size $s$}                                                                                             \\ \cline{2-8} 
                         & 2              & 4              & 8              & 16             & 32             & 64             & 128            \\ \hline
2                        & (0.633, 0.291) & (0.678, 0.272) & (0.719, 0.217) & \textbf{(0.725, 0.145)} & (0.626, 0.100) & (0.722, 0.174) & (0.674, 0.195) \\ \hline
4                        & (0.847, 0.294) & (0.908, 0.277) & (0.886, 0.219) & (0.909, 0.151) & (0.892, 0.103) & \textbf{(0.929, 0.181)} & (0.924, 0.197) \\ \hline
8                        & (0.941, 0.292) & (0.972, 0.273) & (0.971, 0.209) & (0.937, 0.144) & (0.972, 0.100) & (0.991, 0.183) & \textbf{(0.993, 0.192)} \\ \hline
\end{tabular}
\end{subtable}
\end{table*}

\begin{figure*}[t]
\centering
\begin{subfigure}{0.38\linewidth}
\includegraphics[width=\linewidth]{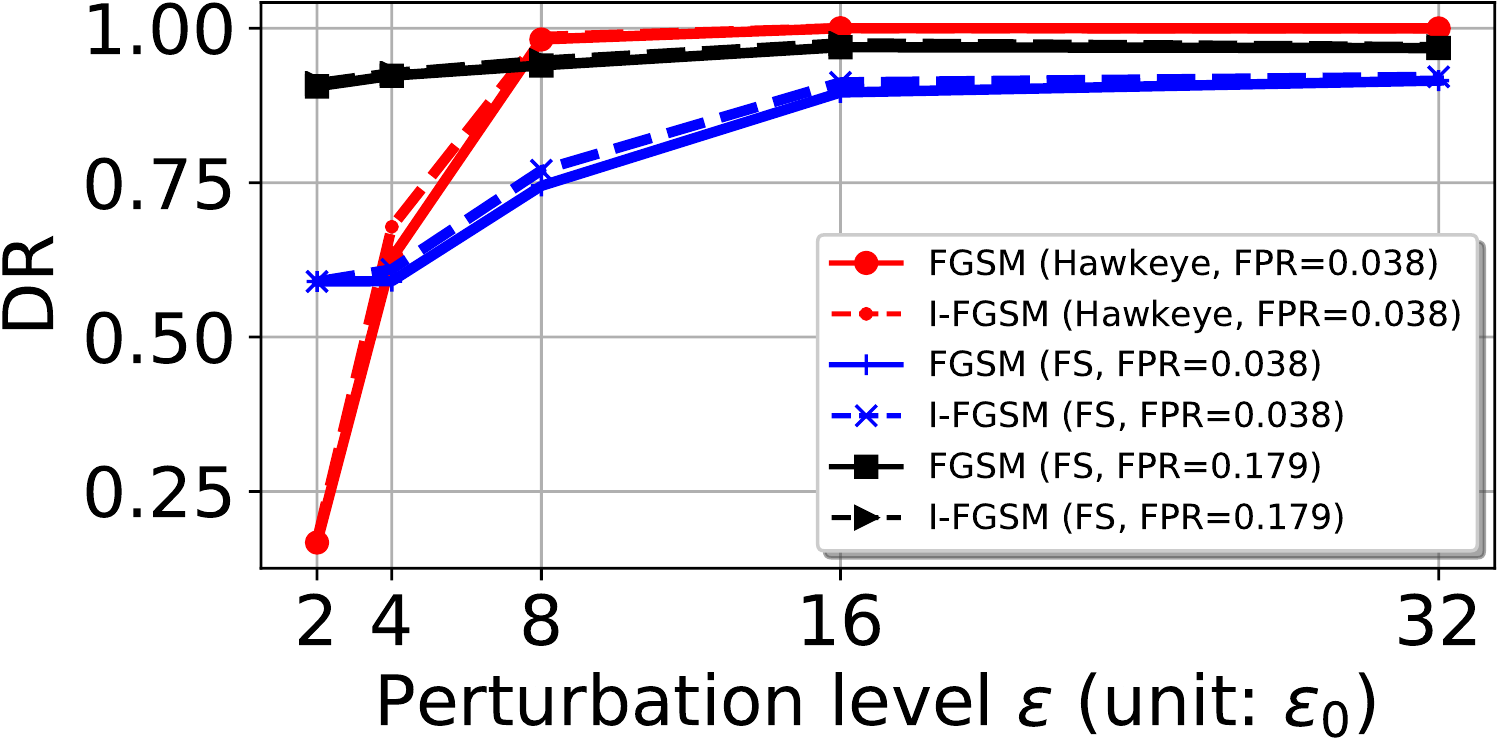}	
\caption{Detection rates in MNIST.} \label{fig:detection_mnist_white}
\end{subfigure}	
\hfil 
\begin{subfigure}{0.38\linewidth}
\includegraphics[width=\linewidth]{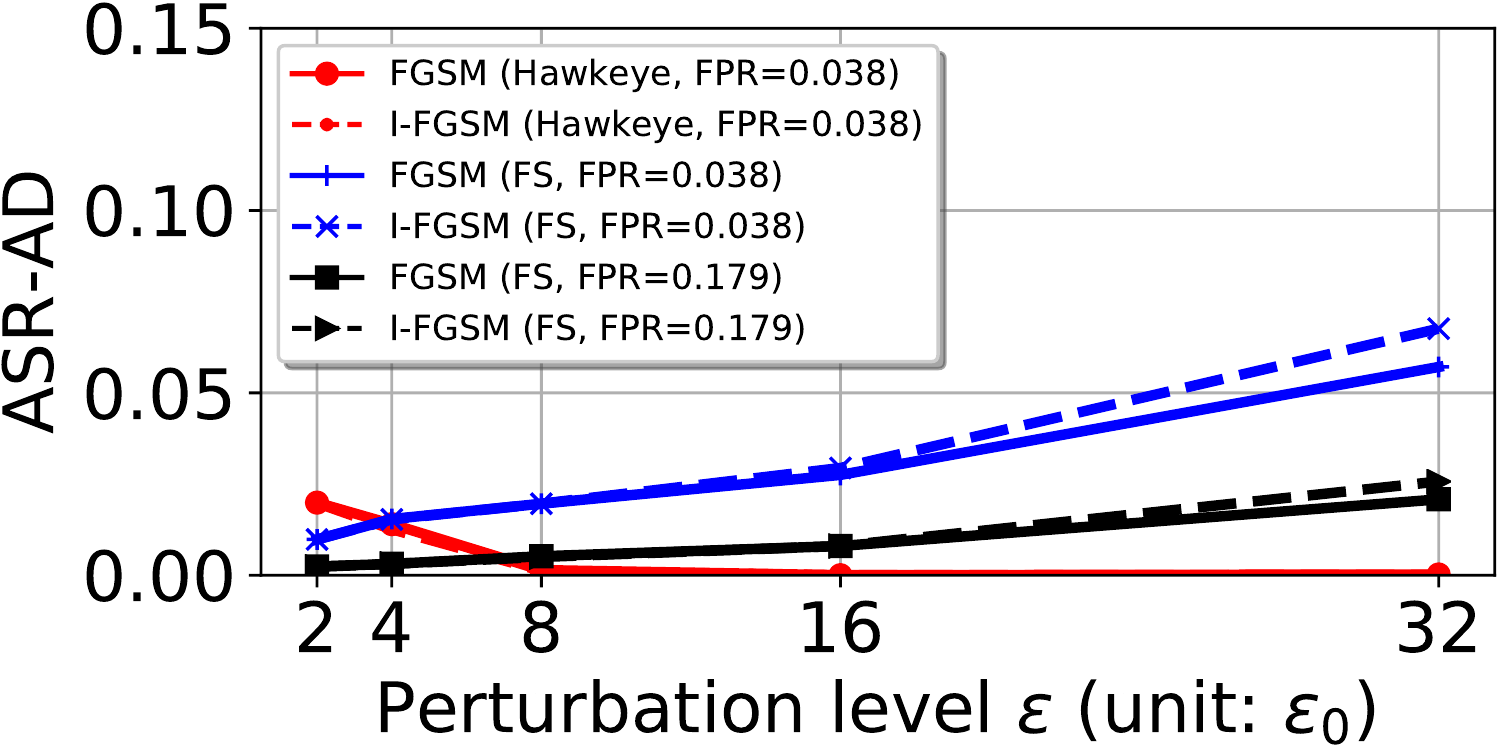}	
\caption{Attack success rates after detection in MNIST.} \label{fig:after_success_mnist_white}
\end{subfigure}	
\\
\begin{subfigure}{0.38\linewidth}
\includegraphics[width=\linewidth]{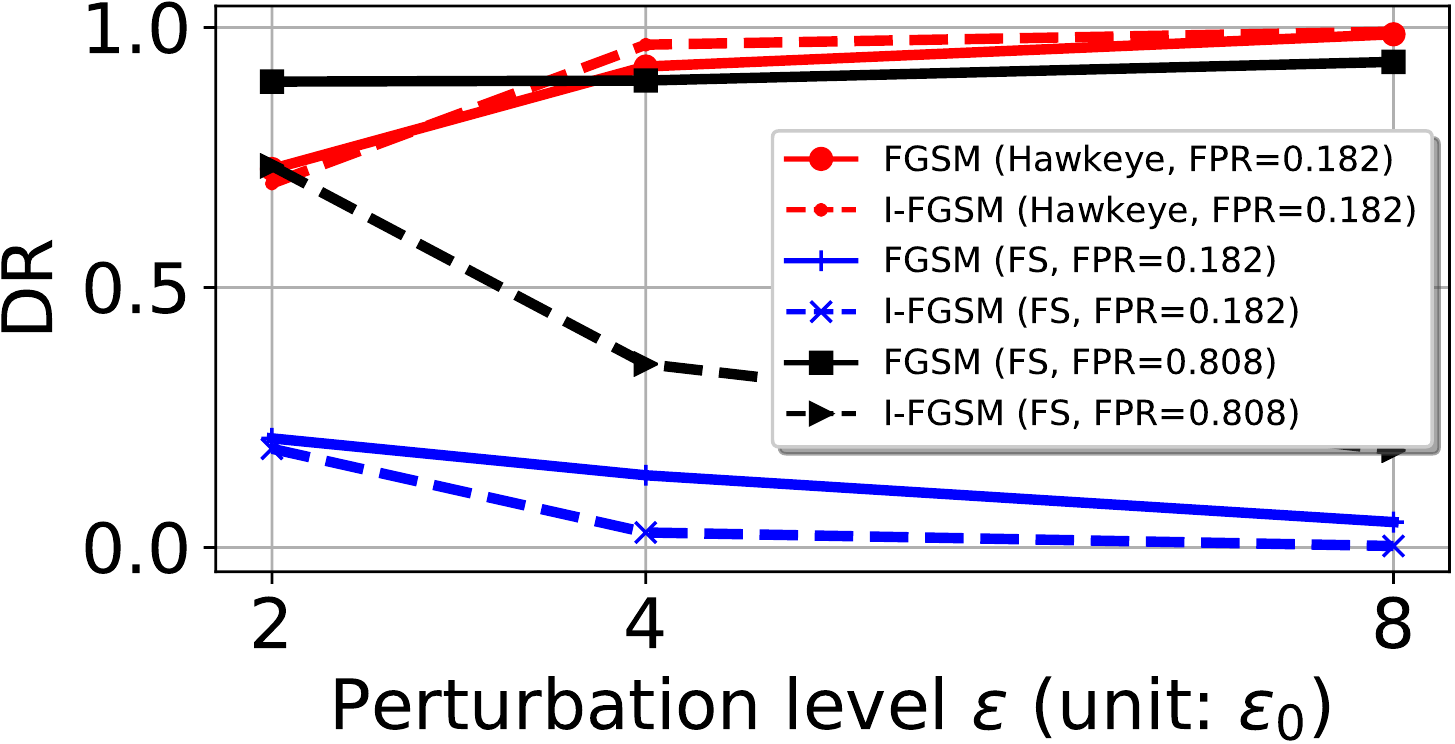}
\caption{Detection rates in ImageNet.} \label{fig:detection_imagenet_white}
\end{subfigure}	
\hfil
\begin{subfigure}{0.38\linewidth}
\includegraphics[width=\linewidth]{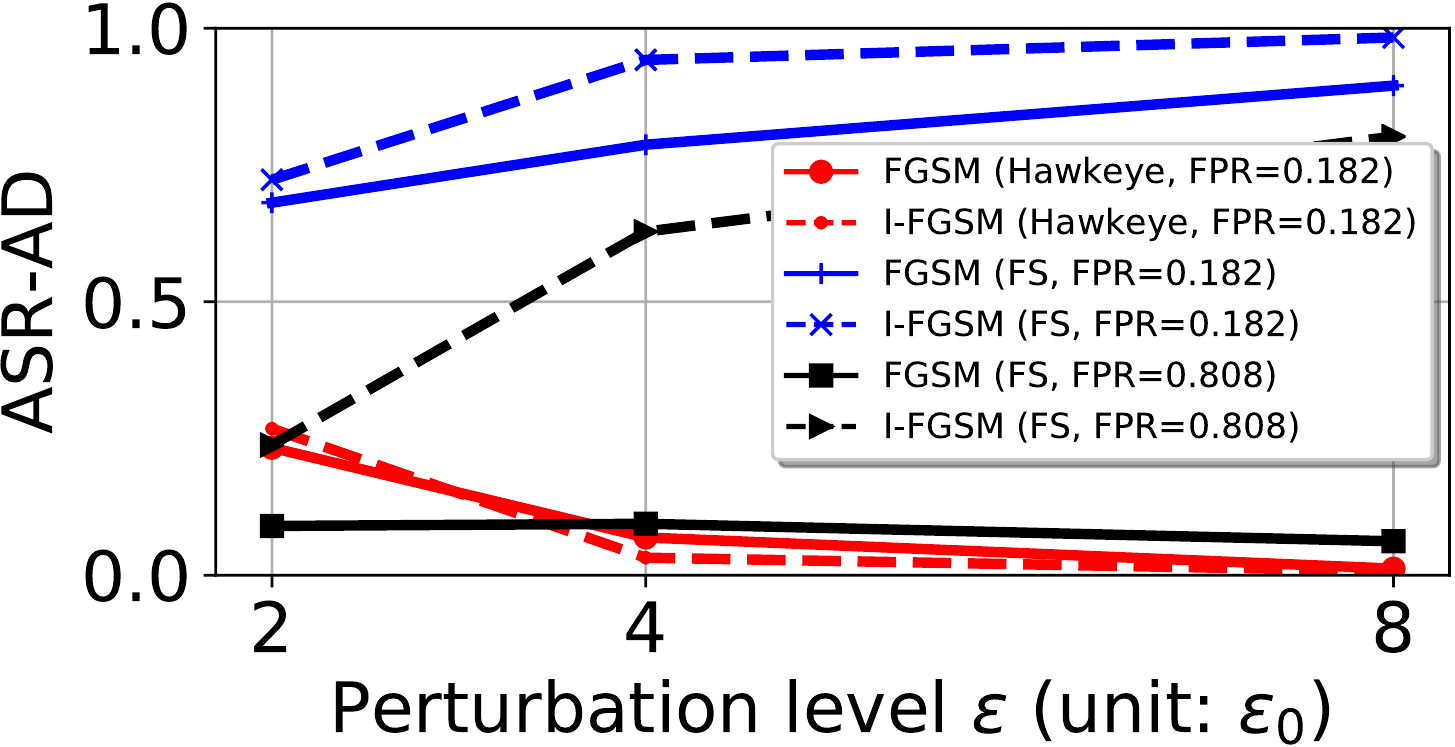}
\caption{Attack success rates after detection in ImageNet.} \label{fig:after_success_imagenet_white}
\end{subfigure}	
\caption{Detection rates and attack success rates after detection.}
\label{fig:rates_white}
\end{figure*}

\begin{figure*}[t]
\centering
\begin{subfigure}{0.38\linewidth}
\includegraphics[width=\linewidth]{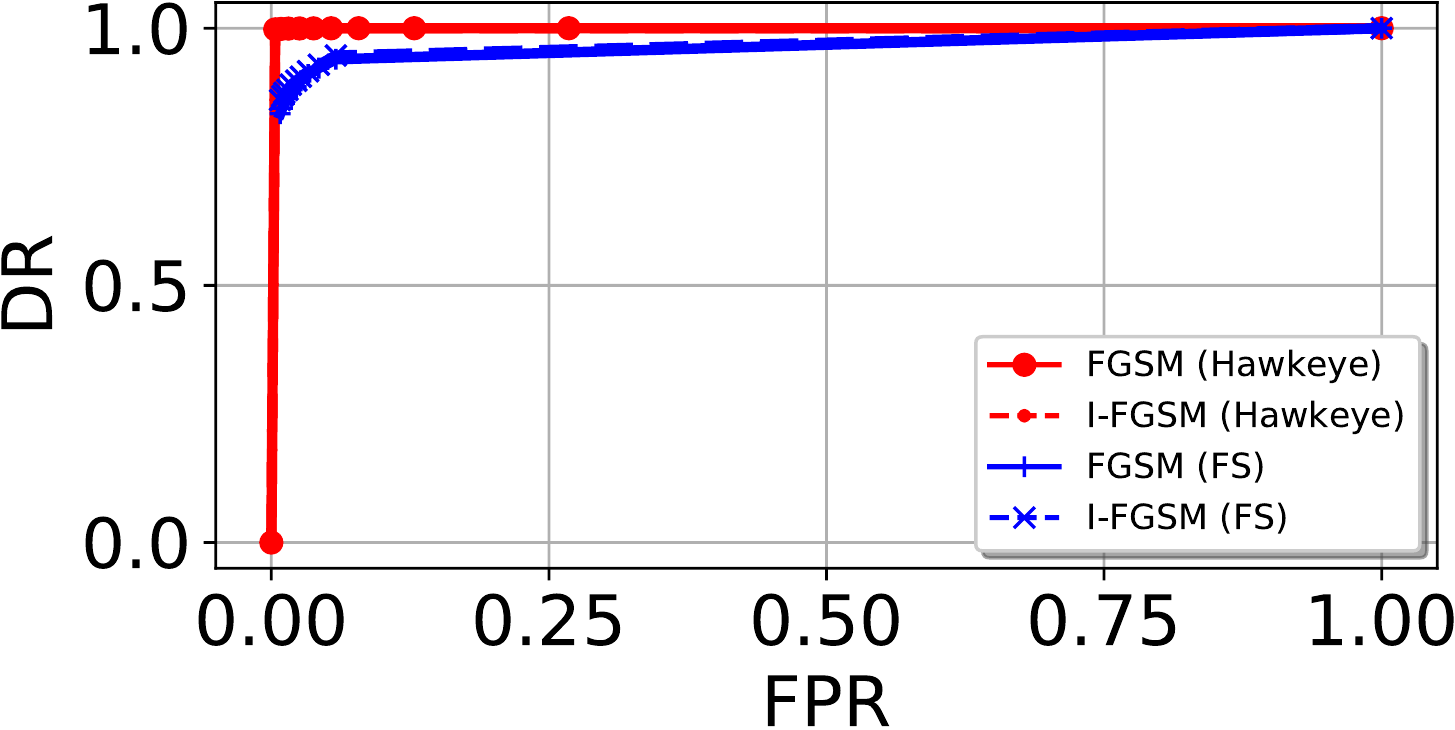}	
\caption{ROC in MNIST when $\epsilon=32\epsilon_0$.} \label{fig:roc_mnist_white}
\end{subfigure}	
\hfil 
\begin{subfigure}{0.38\linewidth}
\includegraphics[width=\linewidth]{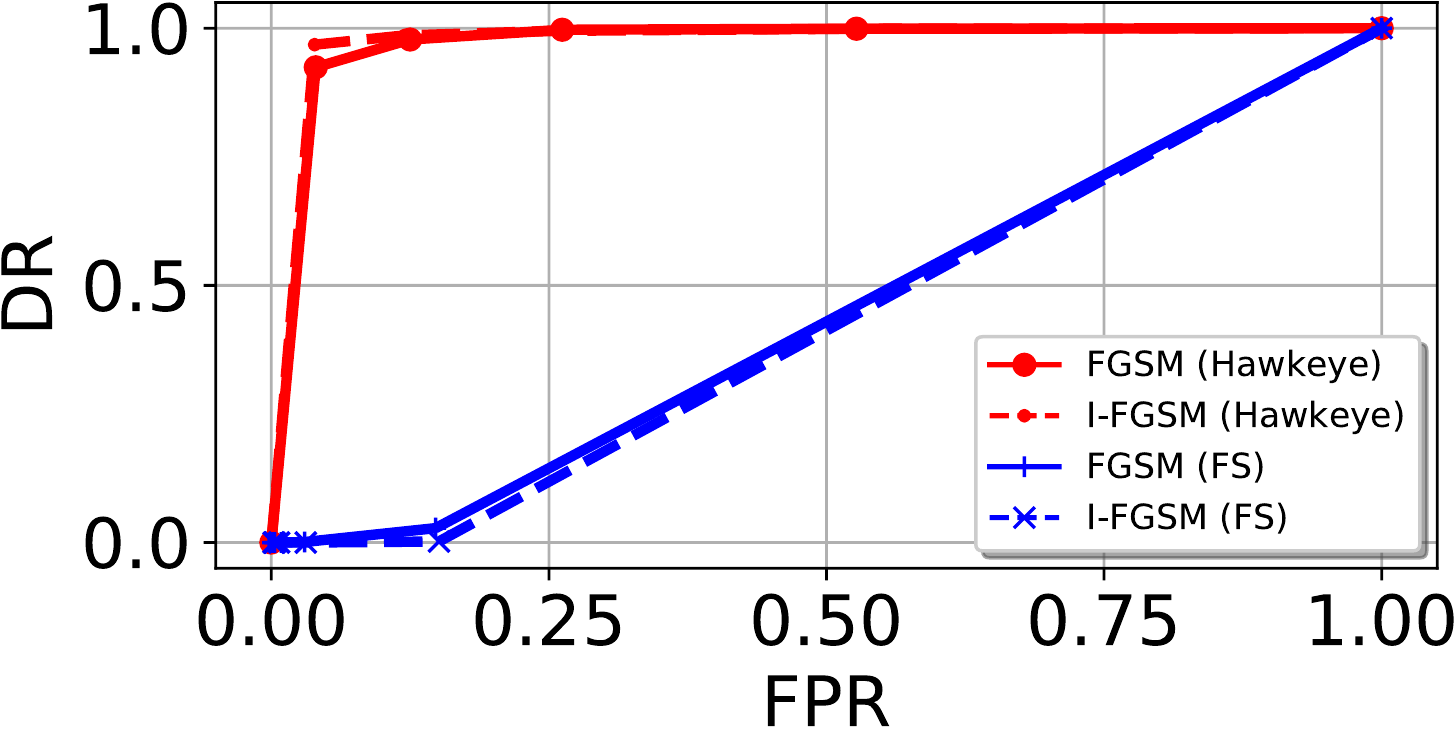}	
\caption{ROC in ImageNet when $\epsilon=8\epsilon_0$.} \label{fig:roc_imagenet_white}
\end{subfigure}	
\caption{Receiver operating characteristic (ROC) curve for \name and Feature Squeezing.}
\label{fig:roc_white}
\end{figure*}


\begin{figure*}[t]
\centering
\begin{subfigure}{0.305\linewidth}
\includegraphics[width=\linewidth]{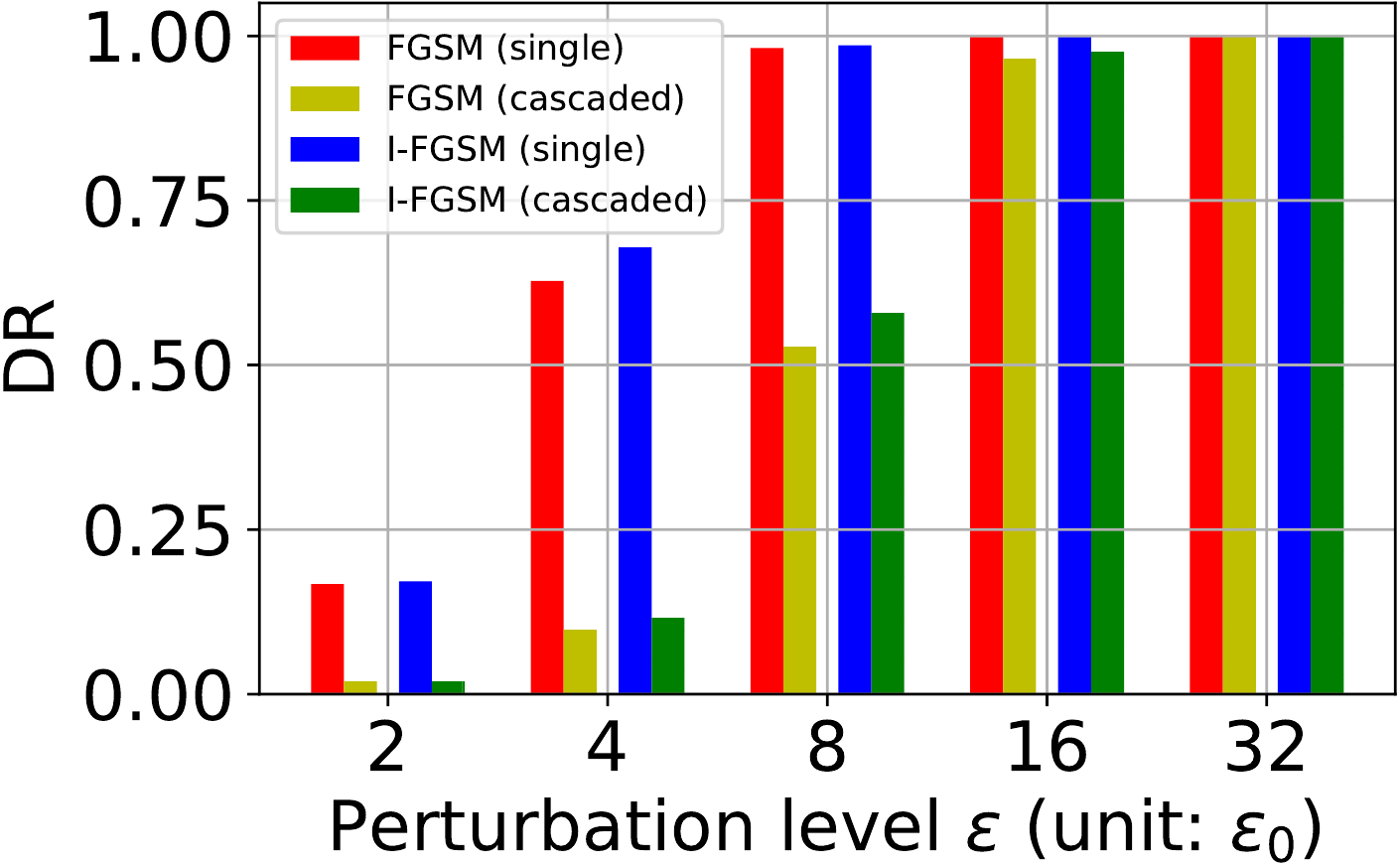}	
\caption{Detection rate comparison in MNIST. Here, the single AED is of FPR$=0.038$ and cascading AEDs achieves FPR$=0.002$.} \label{fig:detection_mnist_multistep_white}
\end{subfigure}	
\hfil 
\begin{subfigure}{0.38\linewidth}
\includegraphics[width=\linewidth]{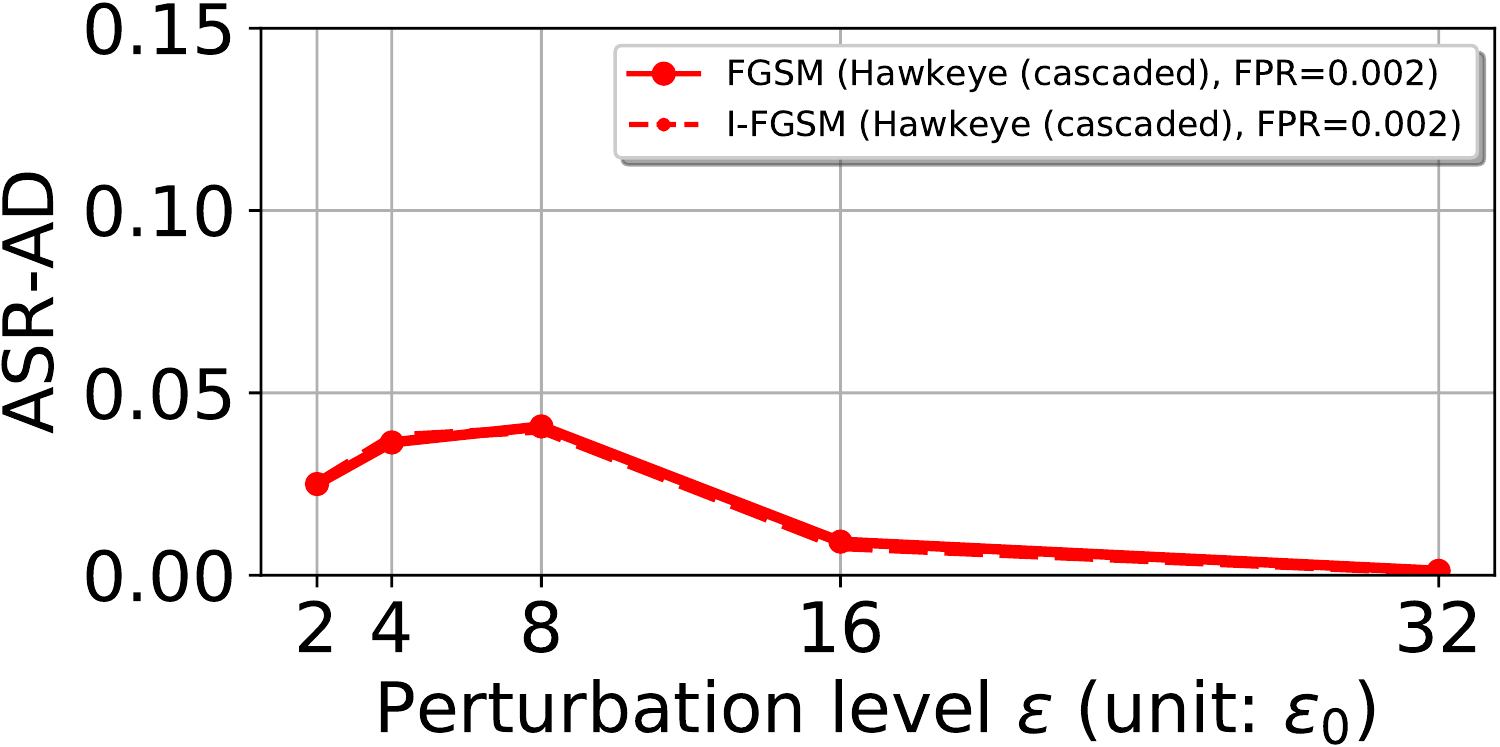}	
\caption{Attack success rates after detection in MNIST.} \label{fig:after_success_mnist_multistep_white}
\end{subfigure}	
\\
\begin{subfigure}{0.305\linewidth}
\includegraphics[width=\linewidth]{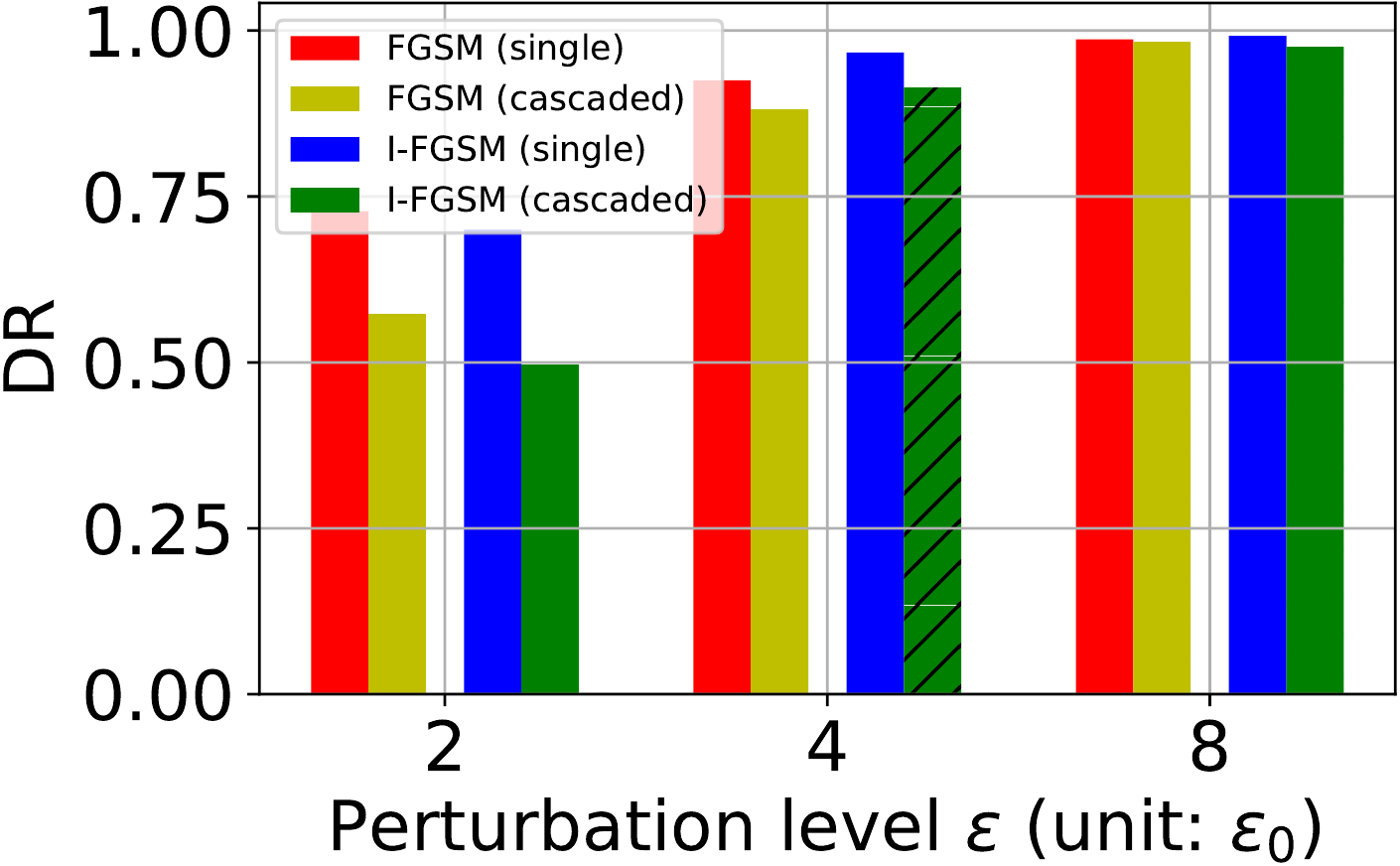}
\caption{Detection rate comparison in ImageNet. Here, the single AED is of FPR$=0.182$ and cascading AEDs achieves FPR$=0.088$.} \label{fig:detection_imagenet_multistep_white}
\end{subfigure}	
\hfil
\begin{subfigure}{0.38\linewidth}
\includegraphics[width=\linewidth]{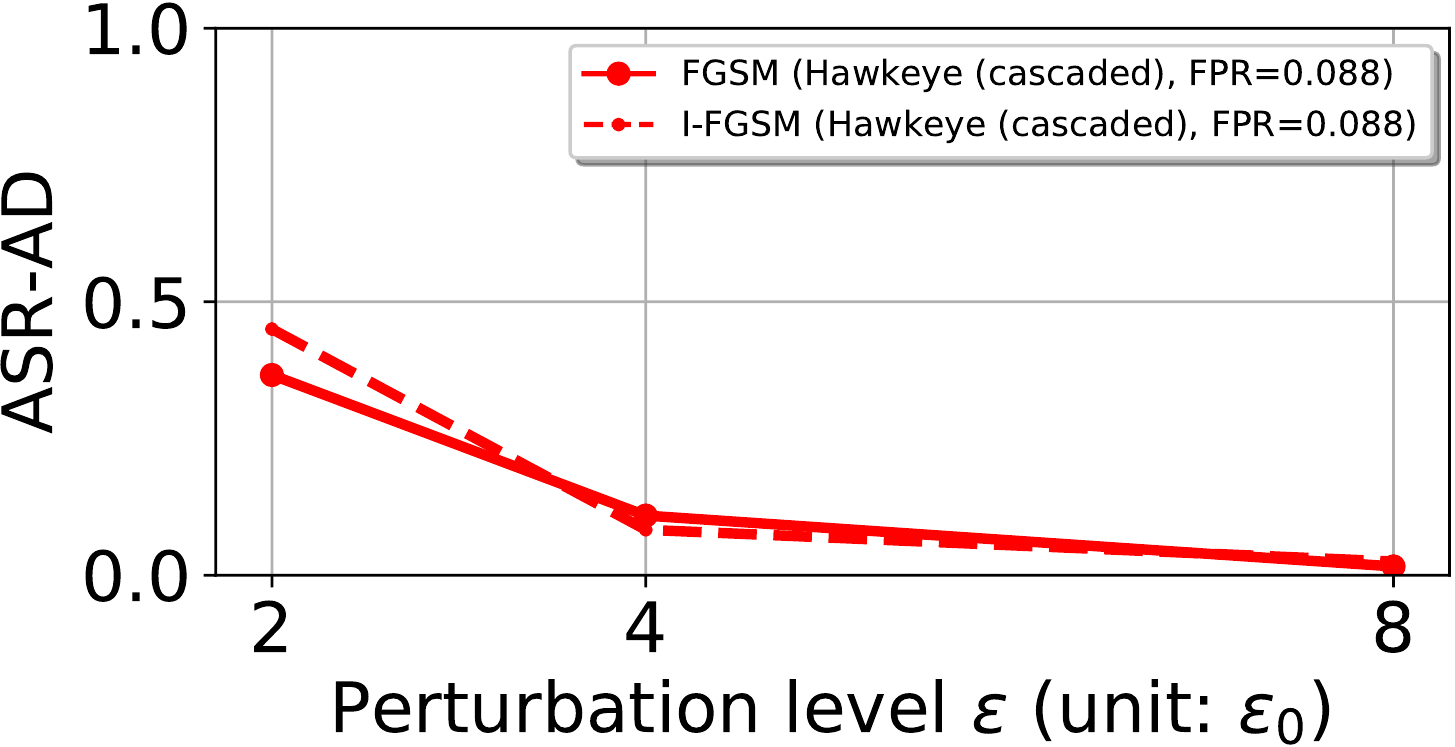}
\caption{Attack success rates after detection in ImageNet.} \label{fig:after_success_imagenet_multistep_white}
\end{subfigure}	
\caption{Detection rates and attack success rates after detection of \name cascading two steps of AEDs.}
\label{fig:rates_multistep_white}
\end{figure*}

\begin{figure*}[t]
\centering
\begin{subfigure}{0.38\linewidth}
\includegraphics[width=\linewidth]{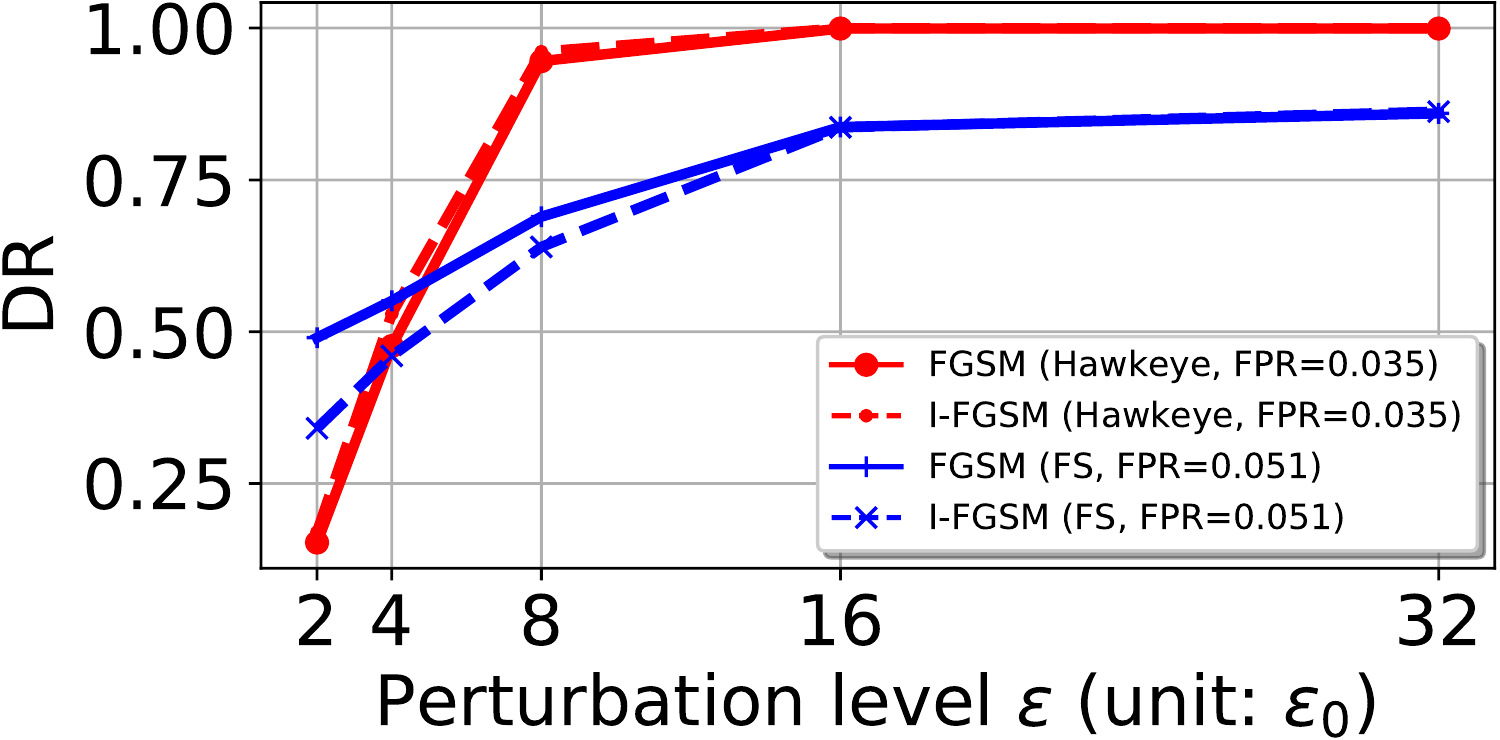}	
\caption{Detection rates in MNIST.} \label{fig:detection_mnist_black}
\end{subfigure}	
\hfil 
\begin{subfigure}{0.38\linewidth}
\includegraphics[width=\linewidth]{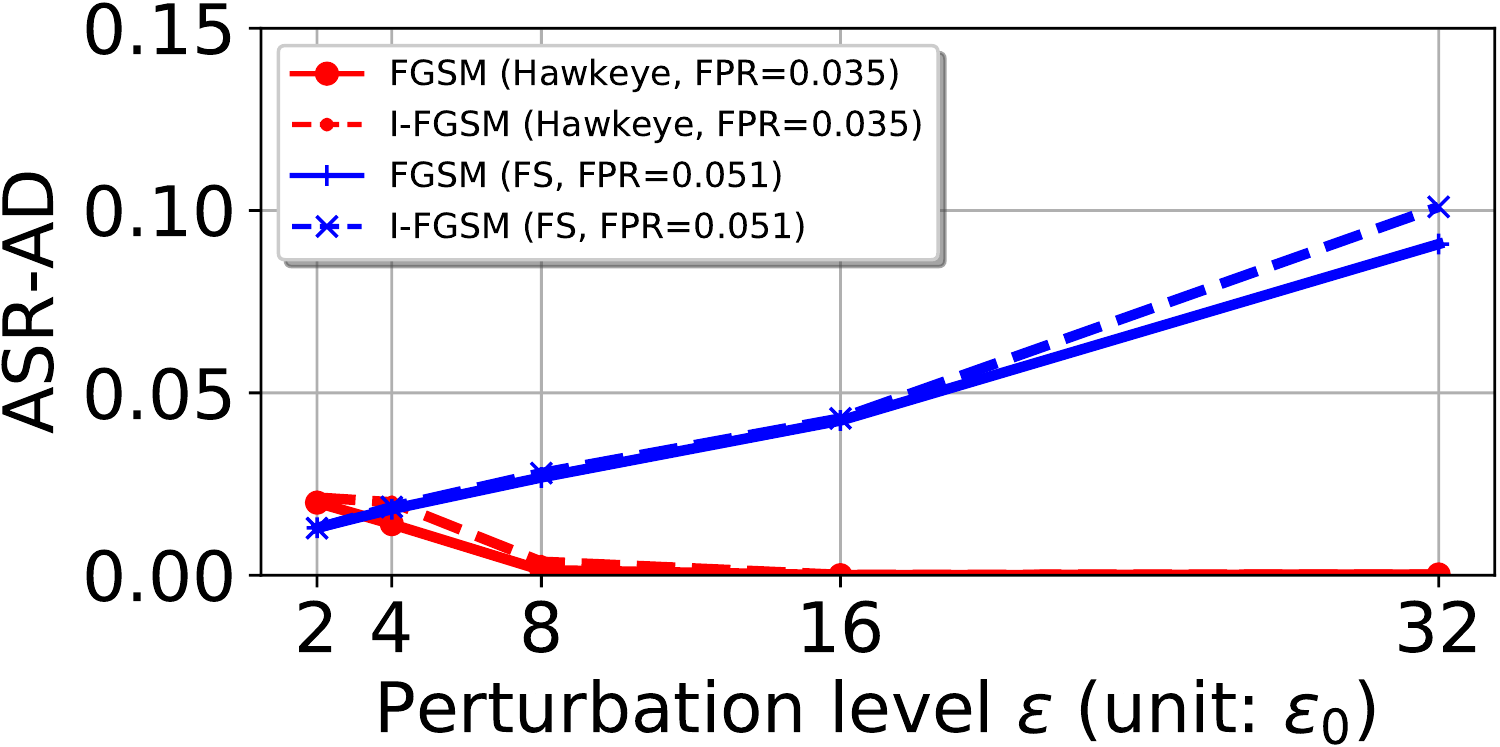}	
\caption{Attack success rates after detection in MNIST.} \label{fig:after_success_mnist_black}
\end{subfigure}	
\\
\begin{subfigure}{0.38\linewidth}
\includegraphics[width=\linewidth]{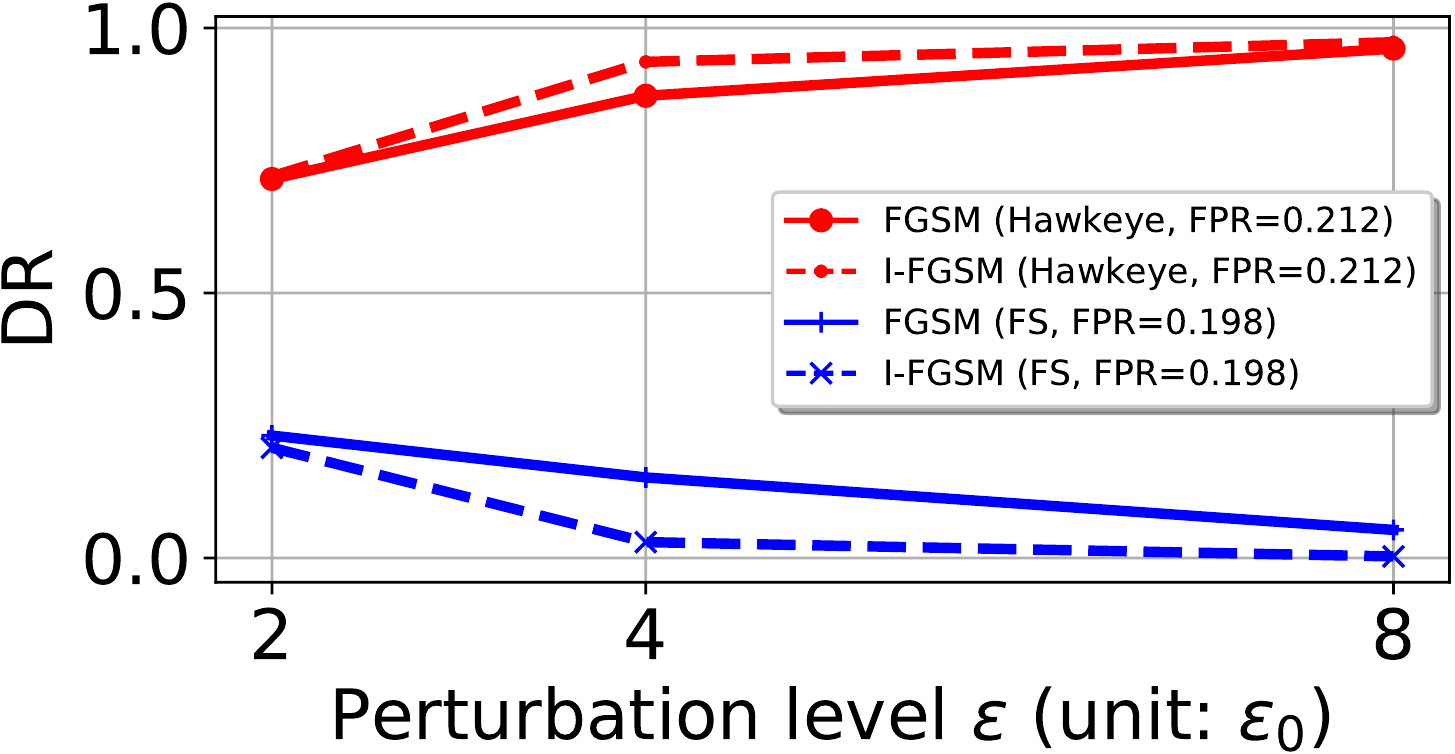}
\caption{Detection rates in ImageNet.} \label{fig:detection_imagenet_black}
\end{subfigure}	
\hfil
\begin{subfigure}{0.38\linewidth}
\includegraphics[width=\linewidth]{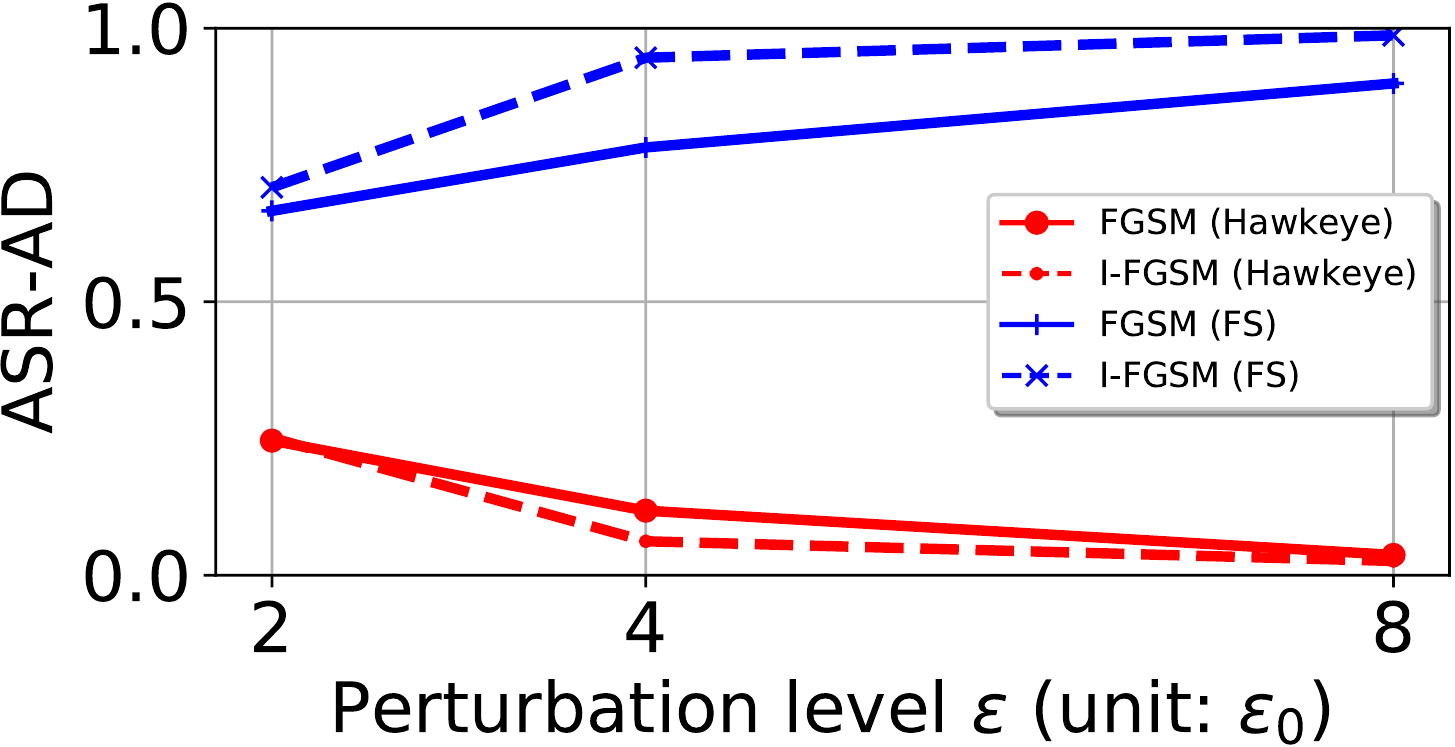}
\caption{Attack success rates after detection in ImageNet.} \label{fig:after_success_imagenet_black}
\end{subfigure}	
\caption{Detection rates and attack success rates after detection when AEs are crafted on a substitute model (\textit{i.e.}, black-box attacks).}
\label{fig:rates_black}
\end{figure*}

We considered four methods for generating AEs. Two are FGSM and I-FGSM discussed in Section \ref{sec:background} and other two are Jacobian-based saliency-map approach (JSMA) \cite{Papernot2016} and Deepfool \cite{Moosavi-Dezfooli15}, which will be described in Section \ref{sec:related} later as related work. However, JSMA was not effective in generating AEs at the same perturbation levels as FGSM or I-FGSM, and it took too long time and too much memory as said in \cite{ndss/Xu0Q18}. Deepfool was good at generating AEs, but  as reported in \cite{ndss/Xu0Q18}, the generated AEs were far too distorted from their original images and thus they can be easily detected by humans. For this reason, we dropped JSMA and Deepfool and use FGSM and I-FGSM in our evaluations. 

For the selected attack methods, Figure \ref{fig:success} shows the attack success rate (ASR) at a given perturbation level $\epsilon$, when the DNN is used without any defense mechanism. ASR is the ratio of the perturbed images that are misclassified by a DNN model to the total number of images being attempted.
In case of ImageNet, we compute the ASR, being associated with top-5 accuracy. That is, we regard the perturbed image an AE when none of the top 5 predictions is the true label. This is because with the top-1 accuracy in ImageNet, the ASR is meaninglessly high~\cite{Kurakin2016b}. The figure shows that I-FGSM is in general a stronger attack than FGSM. We can also see that ImageNet is much more sensitive to AEs than MNIST, showing that I-FGSM can reach almost 100\% ASR even at the perturbation level of only $\epsilon=8\epsilon_0$.

\subsection{AE detector performance according to the step size}

In order to choose the best value of the quantization step size, we train the AED varying the step size $s$, and evaluate the performance at multiple perturbation levels. Table \ref{table:detector_step} shows the results of such experiments in terms of (DR, FPR) when our decision threshold $T$ is 0.5.
When training the AED, we set $\epsilon_{max}$ as 32$\epsilon_0$ for MNIST and 8$\epsilon_0$ for ImageNet. As we can see from Figure \ref{fig:example}, AEs are usually recognizable by humans before the perturbation level reaches $\epsilon_{max}$.

From the table, we can first notice that in both MNIST and ImageNet, the best DR is achieved only when the step size is larger than the perturbation level being tested.
This supports our intuition that quantization process may nullify the malicious noise that is smaller than the quantization step size, and thus the quantized input $x_q$ can be an invariant reference of an image. When the step size is smaller than or comparable to the perturbation level, the quantized input $x_q$ may not serve as a good reference, and thus, the AED experiences the difficulty in distinguishing between normal images and AEs. However, we can also see that too large step size compared to the perturbation level can decrease the DR. Therefore, one specific value for the step size cannot achieve the best performance for all the range of perturbation levels.
Thus, in practice, we suggest to use the step size that performs well for the large perturbation levels (\textit{e.g.}, $s=64$ or $s=128$ in our evaluations). The rationale behind this is that for the low perturbation levels, the ASR for the DNN without any defense is also low and thus after AE detection, the success rates of attacks may become much lower even though the DR of the AED is not that high (see Figure \ref{fig:rates_white}). Further, our AED can be used together with existing defenses that hide the gradients of DNN models, which protect the DNN well for low perturbation levels \cite{Kurakin2016b}.
For this reason, we will focus on studying the performance of our AED using step size $s=64$ hereafter, unless otherwise stated.

\subsection{Performance comparison with Feature Squeezing}

Figure \ref{fig:rates_white} shows the comparison results between \name and FS, the state-of-the-art detector for AE from 2018~\cite{ndss/Xu0Q18}. For this experiment, we use the step size $s=64$ and thus, as shown in Table \ref{table:detector_step}, \name has FPR$=0.038$ for MNIST and FPR$=0.182$ for ImageNet. For fair comparison, we found two decision thresholds for FS in each dataset: one that achieves the same FPR as \name and the other that achieves a comparably high DR. From Figures \ref{fig:detection_mnist_white} and \ref{fig:detection_imagenet_white}, we can see that FS achieves a significantly lower DR than \name at the same FPR. If we try the other decision threshold for FS to achieve the high DR, then FS's FPR goes up to FPR$=0.179$ for MNIST and FPR$=0.808$ for ImageNet, both of which are unacceptably high.
The receiver operating characteristic (ROC) curve in Figure \ref{fig:roc_white} shows that in general, \name outperforms FS in such a way that \name achieves a higher DR at the same FPR, or it achieves a lower FPR at the same DR. The difference is much more significant for the larger and more challenging ImageNet.
The area under the curve (AUC) for ROC curves is 0.99 for \name and 0.96 for FS in MNIST, and 0.97 for \name and 0.43 for FS in ImageNet.

Figures \ref{fig:after_success_mnist_white} and \ref{fig:after_success_imagenet_white} show the attack success rates after detection (ASR-AD), which is the ratio of maliciously perturbed images that pass through a detector without being detected as AE {\em and} mislead a DNN model, to the total number of images being attempted.
One may argue that this measure is more important than DR because these are the examples that both escape detection and finally cause an error in the application DNN. 
We can first see that in general, \name achieves a much lower ASR-AD than FS. Second, we can also see that even though the DR of \name at the low perturbation level is not as high as the DR at the high perturbation level, but it is not a big issue in terms of ASR-AD. Since ASR itself is low at the low perturbation level, not many AEs can be generated at this level. Thus, even with DR that is much lower than 1, the ASR-AD can still be quite low.

\subsection{Performance of cascading AE detectors}\label{sec:eval_cascading}
 
We evaluate \name when we cascade two AEDs of $s=64$ and $s=128$, and Figure \ref{fig:rates_multistep_white} shows the results. As we have seen in Table \ref{table:detector_step}, the two step sizes $s=64$ and $s=128$ achieve a very high DR that is 1 or almost close to 1 at the high perturbation level. Thus, cascading AEDs can still maintain the high DR. We can see that cascading results in an FPR that is significantly lower, as intended. In MNIST, the FPR is 0.002, which was 0.038 for a single AED of $s=64$; thus the 2-cascade achieves an FPR reduction of about 20X. In ImageNet, we now have FPR $=0.088$, which was 0.182 for a single AED of $s=64$, a reduction of about 2X.

On the flip side, cascading AEDs causes a DR at the low perturbation level to be too low. In MNIST, this is not a big issue, since the ASR is also very low. We can see that ASR-AD in MNIST is bounded below 0.04 over all perturbation levels. However, in ImageNet, we can see that the ASR-AD at the perturbation level $\epsilon=2\epsilon_0$ is around 0.4, which is unacceptably high. This is because in ImageNet, the ASR is already above 0.85 at $\epsilon=2\epsilon_0$ as seen in Figure \ref{fig:success}. This result suggests that in ImageNet, \name should be used in combination with another defense mechanism that can protect at low perturbation levels. For example, adversarial training \cite{Kurakin2016a,Kurakin2016b} is especially effective at low perturbation levels. It can control the ASR at $\epsilon=2\epsilon_0$ to be much less 0.1 in ImageNet \cite{Kurakin2016b}.
Thus, when combined with adversarial training, cascaded AEDs in \name will achieve a consistently low ASR-AD over the entire range of perturbation levels, while it can significantly lower FPR compared to a single AED.

\subsection{Performance comparison with Feature Squeezing for black-box attacks}\label{sec:black-box}

So far we have seen the performance of \name when AEs are generated using the target DNN model, \textit{i.e}, white-box attacks. Now, we evaluate how \name performs in comparison with FS when AEs are crafted using a substitute model, \textit{i.e.}, black-box attacks.
As a substitute model, we consider the same DNN model that is trained with different initialization. Existing defenses that hide the gradient of the target DNN are typically less robust against such a substitute model, especially when I-FGSM is used \cite{Tramr17, Kurakin2016b}. For evaluation, we used the detectors that have FPR$=0.038$ for MNIST and FPR$=0.182$ for ImageNet when with white-box attacks in Figure \ref{fig:rates_white}.
The results of our experiment are shown in Figure \ref{fig:rates_black}. We first notice that at many perturbation levels, DR and ASR-AD for both \name and FS become worse than in Figure \ref{fig:rates_white}, but the difference is not significant.
For example, \name's DR at $\epsilon=8\epsilon_0$ in ImageNet was 0.987 at white-box attacks and here, it is 0.961, which is only about 2\% decrement.
We also notice that an FPR may increase when black-box attacks are performed, but not significantly. As an example, black-box attacks cause \name's FPR in ImageNet to increase to 0.212 from 0.182, which is 3\% difference.
We think that this is the virtue of AE detection approach as a defense. Unlike defense mechanisms based on gradient masking, which are vulnerable to black-box attacks,
as has been repeatedly shown \cite{corr/MengC17}
defenses by detecting AEs are robust against such black-box attacks. \name, being a defense mechanism in this category (detecting AEs), possesses this benefit as does FS.

\subsection{Detection of environmental effects}\label{sec:effect}
\begin{figure}[t]
\centering
\begin{subfigure}{0.9\linewidth}
\includegraphics[width=\linewidth]{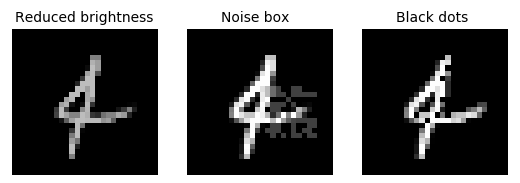}	
\caption{MNIST.} \label{fig:effect_mnist}
\end{subfigure}	
\\
\begin{subfigure}{0.9\linewidth}
\includegraphics[width=\linewidth]{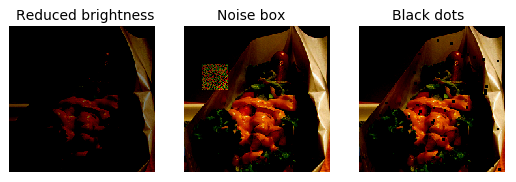}
\caption{ImageNet.} \label{fig:effect_imagenet}
\end{subfigure}	
\caption{An example of environmental effects on camera inputs. When a DNN classifier is used for autonomous vehicles, these can physically occur to a camera. These are not AEs by definition, but can mislead a DNN, effectively working as if AEs.}
\label{fig:effect}
\end{figure}

\begin{table}[t]
\centering
\caption{
Detection rate of \name for environmental effects on camera inputs illustrated in Figure \ref{fig:effect}.
}
\label{table:effect}
\begin{tabular}{cccc}
\hline
         & \begin{tabular}[c]{@{}c@{}}Reduced\\ brightness\end{tabular} & \begin{tabular}[c]{@{}c@{}}Noise\\ box\end{tabular} & \begin{tabular}[c]{@{}c@{}}Black\\ dots\end{tabular} \\ \hline
MNIST    & 0.9866                                                       & 0.9727                                              & 0.9489                                               \\ \hline
ImageNet & 0.822                                                        & 0.856                                               & 0.778                                                \\ \hline
\end{tabular}
\end{table}

We now see if \name can perform well against environmental effects that can physically happen to a camera that creates input images to a DNN classifier, such as in autonomous vehicles. As introduced in \cite{corr/PeiCYJ17}, we consider three effects, brightness-reduced images (simulating images taken at night time), occlusion by a noise box (simulating an attacker or a water drop potentially blocking some parts of a camera), and occlusion by multiple tiny black dots (simulating dirt on camera lens). An example of each effect is shown in Figure \ref{fig:effect}. These are not AEs by definition, but can mislead a DNN prediction by the distortion on images that functions effectively in a similar way to AEs. The reduced brightness is created by deceasing all pixel values by $l$. The noise box effect is made by adding $n\times n$ box at a random position and the noise box is filled with random values over $[-l,l]$. The black dots are made by adding $w\times w$ box whose pixels are all zero at a random position $l$ times.

Figure \ref{fig:effect} is created with $(l,n,w)=(64,10,1)$ in MNIST and with $(l,n,w)=(64,40,4)$ in ImageNet. Table \ref{table:effect} shows the DR of \name in such conditions. We can see that \name achieves a DR of around 0.95 for MNIST and a DR of around 0.8 for ImageNet. Although \name is not trained for such types of images, it can still detect them well. This implies that \name can work well against other potential AE generation methods as well as it can be used for detecting such environmentally distorted images.

\section{Discussion}\label{sec:discussion}

Our ResNet18 model used in ImageNet achieves top-1 and top-5 accuracies as 0.66 and 0.84, respectively, as shown in Table \ref{table:model}. These are slightly lower than those reported in \cite{resnet}, where the top-1 accuracy$= 0.69$ and the top-5 accuracy$=0.89$. This difference mainly occurs because in \cite{resnet}, the accuracies are computed by 10-crop testing that takes 10 different crops from an image and classification scores are then combined to determine the most likely class. The improvement in accuracies actually reduces ASR and ASR-AD. Thus, if we use a better DNN model, like a better-tuned ResNet-18 or deeper ResNets such as ResNet-152, then \name's ASR-AD is expected to be further improved.

\section{Related work}\label{sec:related}

\subsection{Other methods for creating adversarial examples}

Jacobian-based saliency-map approach (JSMA) by \cite{Papernot2016}  aims at fooling a classification model into outputting a specific target class $t$ by iteratively perturbing one or two pixels at a time.
Since only one or two pixels change at a time, this method takes a long time to create an AE  (unusably long for ImageNet images) compared to FGSM or I-FGSM \cite{corr/abs-1712-07107,ndss/Xu0Q18}, and the resulting AEs usually contain pixels with high intensity, which can be relatively easily detected by humans. 
JSMA also requires a huge amount of memory to compute a Jacobian matrix \cite{ndss/Xu0Q18}. For these reasons, we did not consider it in our evaluation.

Deepfool \cite{Moosavi-Dezfooli15} is another untargeted attack method, creating AEs with the assumption that a classifier is a linear model. Since DNNs are not actually linear, Deepfool iterates a process, where a DNN is first approximated to a linear model and the minimum level of perturbation is decided for each approximation. Although Deepfool can successfully mislead DNNs, AEs created by it often look too distorted from their original images, thus easily detectable by humans, as reported in \cite{ndss/Xu0Q18}. Thus, we exclude it from our evaluation.

\subsection{Defenses by hiding gradients}\label{sec:existing_defenses}
The basis of the first generation defense mechanisms against adversarial examples is what is often called gradient masking, which attempts to hide a useful gradient (like $\frac{\partial J(x,y)}{\partial x}$ or $\frac{\partial [F(x)]_t}{\partial x}$) in the vicinity of the input data points. Below are representative defenses of such a kind.

Contractive penalty was considered in \cite{Gu14}, where as a regularizer, $\lVert \frac{\partial F(x)}{\partial x} \rVert_2$ is added into the training cost. This is intended to train a model, minimizing the norm of Jacobian $\frac{\partial F(x)}{\partial x}$ so that the gradient near input data points is reduced.

Adversarial training by \cite{Goodfellow2015} enhances robustness by exposing a model to AEs in advance during the training phase. This is usually achieved by modifying the cost function to $J_a(x,y)$ as
\begin{align}
J_a(x,y) = &(1-\alpha) J(x,y)\nonumber\\
& + \alpha J\left(x + \epsilon \text{sign}\left(\frac{\partial J(x,y)}{\partial x} \right),y\right),
\label{eq:at}
\end{align}
where $0<\alpha<1$. 
Namely, the second term based on an AE generated by FGSM in \eqref{eq:fgsm} is considered at every step of training as a regularizer.
Adversarial training shows the gradient masking effect \cite{Tramr17}, and also changes the decision boundary by forcing an image $x$ and an adversarial example created from $x$ to the same class.

Distillation is a recent advance in deep learning by \cite{Hinton15}, which found that knowledge in a neural network can be transferred to a smaller model. It was shown in \cite{Papernot_SP16} that a distilled model can be more robust to adversarial examples than its original model, and the robustness gets larger as we use a higher value for the so-called distillation temperature $T$ in the training phase.
However, \cite{Papernot2017} reported later that the distilled model is weak against black-box attacks.

Saturating networks was proposed in \cite{Nayebi17}, which 
encourages activation functions to stay in the saturating regime of their non-linearity (\textit{e.g.}, all-zero region for a Relu) by adding a regularizer in the training cost. This is intended to make a model be less sensitive to input perturbations.
	
The methods described above except adversarial training later turned out to fail against black-box attacks \cite{Papernot2017,openAI}, since the adversary can reconstruct the gradients using a substitute model. Adversarial training is relatively robust against black-box attacks, but its protection capacity drops rapidly with the increase in the perturbation levels \cite{Kurakin2016b}.

\subsection{Defenses by detecting adversarial examples}

Grosse \textit{et al.} \cite{corr/GrosseMP0M17} proposed a statistical test to detect AEs from training dataset using maximum mean discrepancy.
This method requires a large set of normal images and their corresponding AEs, and is not capable of detecting individual AEs. Thus, they also proposed detecting individual AEs by adding an additional class to a DNN model, and train the model to recognize AEs as this new class.
However, it is disadvantageous to require changes to the target DNN itself.

Metzen \textit{et al.} \cite{metzen2017detecting} proposed attaching a CNN-based detector at the middle layers of a DNN model. The detector is trained in a supervised manner to classify an input to a normal image or an AE. This requires a large set of AEs to be generated offline for each perturbation level. Attaching a detector in the middle layers also causes the input size to the detector is huge, thereby requiring the detector itself is also another DNN to fully extract the features from such a large input.

MagNet \cite{corr/MengC17} also used a reference input to detect an AE. It uses an autoencoder and compares the input image with the autoencoder output. Since the autoencoder is supposed to reconstruct a given input to the one that is smoothed over possible additive noise, the output of the autoencoder plays a role as a reference. However, training autoencoder to obtain a faithful reconstruction is a challenging task, especially when the inputs have a wide variety like in ImageNet. For this reason, MagNet falls behind FS in ImageNet \cite{ndss/Xu0Q18}.

\section{Conclusion}\label{sec:concusion}
In this paper, we proposed \name to detect Adversarial Examples (AEs). Unlike the first generation defense methods that apply gradient masking, \name does not prevent AEs from being generated. Rather, \name attempts to detect AEs by considering differences in the logit vector generated by the application DNN for the original image and a reference image generated from it.
The reference image is generated by quantizing the original image and we leverage the fact that the quantized image is less sensitive to adversarial modifications. \name designs a separate neural network, an adversarial example detector (AED), which
can observe the difference between the output of the application DNN with the original image and the quantized image and classify whether the original image is benign or adversarial. 
We also propose cascading our AEDs that are trained with different quantization step sizes.
Cascading AEDs maintains the high DR of a single AED but can reduce FPR dramatically.
\name is complementary to prior work based on gradient masking and can be used together to achieve better protection, especially at low perturbation levels.

Our evaluation results using two widely used datasets, MNIST and ImageNet, showed that compared to Feature Squeezing, which is the state-of-the-art AE detection technique, \name is superior in both DR and FPR, achieving the ROC curves that are positioned higher than those of FS.
Compared to an individual AED, cascading our AEDs of step sizes $s=64$ and $s=128$ could reduce the FPR by about 20X in MNIST and by about 2X in ImageNet.
Experimental results also showed that \name is not sensitive to black-box attacks, achieving DR and FPR only slightly worse than those in white-box attacks.

As future work, we will try to find other reference inputs that may work better than quantization, thereby improving detection performance, especially for the small perturbation levels. Multi-modal inputs, such as dual camera used in autonomous vehicles, are a promising avenue to create a good reference input.

\bibliographystyle{unsrt}
\bibliography{kooj}

\end{document}